\documentclass[10pt,twocolumn,letterpaper]{article}

\usepackage{cvpr, times, graphicx, amsmath, amssymb}
\usepackage{amsthm, verbatim, enumerate, color, url, graphicx, mhsetup, wrapfig}
\usepackage[caption=false]{subfig} 
\usepackage{multirow, paralist, xcolor, comment, algorithm, algorithmic, enumitem, lipsum}
\usepackage[pagebackref=true, breaklinks=true, letterpaper=true, colorlinks, bookmarks=false]{hyperref}

\cvprfinalcopy

\ifcvprfinal\fi
\def\cvprPaperID{1074} 

\theoremstyle{plain}

\theoremstyle{definition}
\newtheorem*{definition}{Definition}  

\newcommand{\mmf}{\mathcal{M}} \newcommand{\graph}{\ensuremath{\mathcal G}} \newcommand{\tgraph}{\ensuremath{\tilde{\mathcal G}}} 
\newcommand{\err}{\ensuremath{\mathcal E}} \newcommand{\Sind}{\ensuremath{\mathcal S}} \newcommand{\ins}{\ensuremath{\mathcal I}} 
\newcommand{\tSind}{\ensuremath{\tilde{\mathcal S}}}  
\newcommand{\ttt}{\ensuremath{\tilde{t}}} \newcommand{\htt}{\ensuremath{\hat{t}}}
\newcommand{\ts}{\ensuremath{\tilde{s}}} \newcommand{\hs}{\ensuremath{\hat{s}}}

\newcommand{\C}{\ensuremath{\mathbf{C}}}  
\newcommand{\tC}{\ensuremath{\tilde{\mathbf{C}}}}

\newcommand{\R}{\ensuremath{\mathbf{\Lambda}}}

\newcommand{\Q}{\ensuremath{\mathbf{ Q}}}

\newcommand{\I}{\ensuremath{\mathbf{I}}} 
\newcommand{\A}{\ensuremath{\mathbf{A}}} 
\newcommand{\B}{\ensuremath{\mathbf{B}}} 
\newcommand{\Orth}{\ensuremath{\mathbf{O}}}

\newcommand{\tm}{\ensuremath{\tilde{m}}} 
  
\newcommand{\uvec}{\ensuremath{{\bf u}}}
\newcommand{\wvec}{\ensuremath{{\bf w}}}

\newcommand{\argmin}{\mathop{\mathrm{argmin}}}
\newcommand{\argmax}{\mathop{\mathrm{argmax}}}

\usepackage{soul}

\begin{document}
\title{The Incremental Multiresolution Matrix Factorization Algorithm}
\author{Vamsi K. Ithapu$^\dag$, Risi Kondor$^\S$, Sterling C. Johnson$^\dag$, Vikas Singh$^\dag$\\
$^\dag$University of Wisconsin-Madison,
$^\S$University of Chicago\\
{\small \url{http://pages.cs.wisc.edu/~vamsi/projects/incmmf.html}}
 }
\maketitle


\begin{abstract}
Multiresolution analysis and matrix factorization are foundational tools in computer vision.
In this work, we study the interface between these two distinct topics and obtain techniques to uncover hierarchical block structure in symmetric matrices
-- an important aspect in the success of many vision problems.
Our new algorithm, the {\it incremental multiresolution matrix factorization}, uncovers such structure one feature at a time, and hence scales well to large matrices.
We describe how this multiscale analysis goes much farther than what a direct ``global'' factorization of the data can identify. 
We evaluate the efficacy of the resulting factorizations for relative leveraging within regression tasks using medical imaging data.  
We also use the factorization on representations learned by popular deep networks, 
providing evidence of their ability to infer semantic relationships even when they are not explicitly trained to do so.
We show that this algorithm can be used as an exploratory tool to improve the network architecture, and within numerous other settings in vision. 
\end{abstract}


\section{Introduction} \label{sec:intro}

Matrix factorization lies at the heart of a spectrum of computer vision problems. 
While the wide ranging and extensive use of factorization schemes within structure from motion \cite{sturm1996factorization}, 
face recognition \cite{turk1991eigenfaces} and motion segmentation \cite{cheriyadat2009non} have been known, in the last decade, there is renewed interest in these ideas.
Specifically, the celebrated work on low rank matrix completion \cite{candes2009exact} has enabled deployments in a broad cross-section of vision problems from
independent components analysis \cite{hyvarinen2013independent} to dimensionality reduction \cite{wright2009robust} to online background estimation \cite{xu2013gosus}. 
Novel extensions based on Robust Principal Components Analysis \cite{de2001robust, candes2009exact} are being developed each year. 

In contrast to factorization methods, 
a distinct and rich body of work based on early work in signal processing is arguably even more extensively utilized in vision. 
Specifically, Wavelets \cite{rubinstein2010dictionaries} and other related ideas (curvelets \cite{candes2000curvelets}, shearlets \cite{kutyniok2012shearlets}) 
that loosely fall under multiresolution analysis (MRA) based approaches drive an overwhelming majority of techniques 
within feature extraction \cite{manjunath1996texture} and representation learning \cite{rubinstein2010dictionaries}.
Also, Wavelets remain the ``go to'' tool for image denoising, compression, inpainting, shape analysis and other applications in video processing \cite{meyer1993wavelets}. 
SIFT features can be thought of as a special case of the so-called Scattering Transform (using theory of Wavelets) \cite{bruna2013invariant}.
Remarkably, the ``network'' perspective of Scattering Transform at least partly explains the invariances being identified by deep representations, 
further expanding the scope of multiresolution approaches informing vision algorithms. 

The foregoing discussion raises the question
of whether there are any interesting bridges between Factorization and Wavelets.
This line of enquiry has recently been studied for the most common ``discrete'' object encountered in vision -- graphs. 
Starting from the seminal work on Diffusion Wavelets \cite{coifman2006diffusion}, 
others have investigated tree-like decompositions on matrices \cite{lee2008treelets}, and organizing them using wavelets \cite{gavish2012sampling}.
While the topic is still nascent (but evolving),
these non-trivial results suggest that the confluence of these seemingly distinct topics potentially holds much promise for vision problems \cite{hwa2016latent}. 
Our focus is to study this interface between Wavelets and Factorization, and demonstrate the immediate set of problems that can potentially benefit.  
In particular, we describe an efficient (incremental) multiresolution matrix factorization algorithm. 

\begin{figure}[!t]\centering
\includegraphics[width=0.45\textwidth]{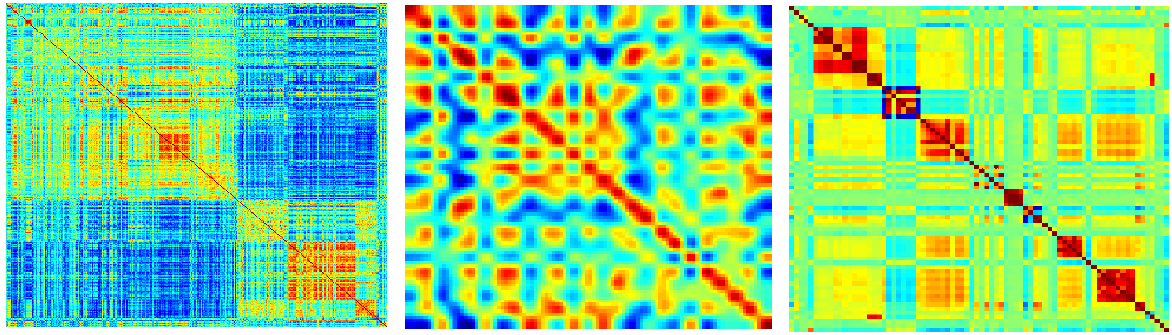}
\caption{\footnotesize \label{fig:covs} Left-to-right example category (or class) covariances from AlexNet, VGG-S (of a few ImageNet classes) and medical imaging data.}
\end{figure}

To concretize the argument above, consider a representative example in vision and machine learning where a factorization approach may be deployed. 
Figure \ref{fig:covs} shows a set of covariance matrices computed from the representations learned by AlexNet \cite{krizhevsky2012imagenet}, 
VGG-S \cite{chatfield2014return} (on some ImageNet classes \cite{russakovsky2015imagenet}) and medical imaging data respectively. 
As a first line of exploration, we may be interested in 
characterizing the apparent parsimonious ``structure'' seen in these matrices.
We can easily verify that invoking the de facto constructs like sparsity, low-rank or a decaying eigen-spectrum cannot account for the ``block'' or cluster-like structures inherent in this data.
Such block-structured kernels were the original motivation for block low-rank and hierarchical factorizations \cite{savas2011clustered, chandrasekaran2005fast} ---
but a multiresolution scheme is much more natural --- in fact, ideal --- if one can decompose the matrix in a way that the blocks automatically `reveal' themselves at multiple resolutions. 
Conceptually, this amounts to a sequential factorization while accounting for the fact that each level of this hierarchy must correspond to approximating some non-trivial structure in the matrix. 
A recent result introduces precisely such a {\it multiresolution matrix factorization} (MMF) algorithm for symmetric matrices \cite{kondor2014}.

Consider a symmetric matrix $\C\in\mathbb{R}^{m \times m}$. 
PCA decomposes $\C$ as $\Q^T\R\Q$ where $\Q$ is an orthogonal matrix, which, in general, is dense. 
On the other hand, sparse PCA (sPCA) \cite{zou2006sparse} imposes sparsity on the columns of $\Q$, allowing for fewer dimensions to interact that may not capture global patterns.
The factorization resulting from such individual low-rank decompositions cannot capture hierarchical relationships among data dimensions.
Instead, MMF applies a sequence of carefully chosen sparse rotations $\Q^1,\Q^2,\ldots \Q^L$ to 
factorize $\C$ in the form
\begin{equation*}
\C = (\Q^1)^T(\Q^2)^T\ldots(\Q^L)^T\R\Q^L\ldots\Q^2\Q^1,  
\end{equation*}
thereby uncovering soft hierarchical organization of different rows/columns of $\C$.
Typically the $\Q^\ell$s are sparse $k^{\textrm{th}}$-order rotations (orthogonal matrices that are the identity except for at most $k$ of their rows/columns), 
leading to a hierarchical tree-like matrix organization. 
MMF was shown to be an efficient compression tool \cite{mmc} and a preconditioner \cite{kondor2014}.
Randomized heuristics have been proposed to handle large matrices \cite{pmmf}.
Nevertheless, factorization involves searching a combinatorial space of row/column indices, which restricts the order of the rotations to be small (typically, $\leq 3$).
Not allowing higher order rotations restricts the richness of the allowable block structure, 
resulting in a hierarchical decomposition that is ``too localized'' to be sensible or informative 
(reverting back to the issues with sPCA and other block low-rank approximations). 

A fundamental property of MMF is the sequential composition of rotations.
In this paper, we exploit the fact that the factorization can be parameterized in terms of an {\it MMF graph} defined on a sequence of higher-order rotations.
Unlike alternate batch-wise approaches \cite{mmc}, we start with a small, randomly chosen block of $\C$, 
and gradually `insert' new rows into the factorization -- hence we refer to this as an {\it incremental} MMF. 
We show that this insertion procedure manipulates the topology of the MMF graph, thereby providing an efficient algorithm for constructing higher order MMFs. 
Our {\bf contributions} are:
{\bf (A)} We present a fast and efficient incremental procedure for constructing higher order (large $k$) MMFs on large dense matrices; 
{\bf (B)} We evaluate the efficacy of the higher order factorizations for relative leveraging of sets of pixels/voxels in regression tasks in vision; and 
{\bf (C)} Using the output structure of incremental MMF, we visualize the semantics of categorical relationships inferred by deep networks, 
and, in turn, present some exploratory tools to adapt and modify the architectures. 


\section{Multiresolution Matrix Factorization} \label{sec:prelim}

\noindent {\bf Notation:} We begin with some notation. Matrices are bold upper case, vectors are bold lower case and scalars are lower case. 
$[m] := \{1,\ldots,m\}$ for any $m \in \mathbb{N}$. 
Given a matrix $\C \in \mathbb{R}^{m \times m}$ and two set of indices 
$\Sind_1=\{r_1,\ldots r_k\}$ and $\Sind_2=\{c_1,\ldots c_p\}$, $\C_{\Sind_1,\Sind_2}$ will denote the block of $\C$ cut out by the rows $\Sind_1$ and columns $\Sind_2$. 
$\C_{:,i}$ is the $i^{th}$ column of $\C$.
$\I_m$ is the $m$-dimensional identity. $SO(m)$ is the group of $m$ dimensional orthogonal matrices with unit determinant.
$\mathcal{R}^m_{\Sind}$ is the set of $m$-dimensional symmetric matrices which are diagonal except for their $\Sind \times \Sind$ block 
($\Sind$--core-diagonal matrices).


Multiresolution matrix factorization (MMF), introduced in \cite{kondor2014, pmmf}, 
retains the locality properties of sPCA while also capturing the global interactions provided by the many variants of PCA, 
by applying not one, but multiple {\it sparse} rotation matrices to $\C$ in sequence. We have the following. 
\begin{definition} \label{thm:mmfdef}
Given an appropriate class $\mathcal{O} \subseteq SO(m)$ of sparse rotation matrices, 
a depth parameter $L \in \mathbb{N}$ and a sequence of integers $m = d_0 \geq d_1 \geq \ldots \geq d_L \geq 1$, 
the {\bf multi-resolution matrix factorization (MMF)} of a symmetric matrix $\C\in\mathbb{R}^{m\times m}$ is 
a factorization of the form 
\begin{equation} \label{eq:mmfdef}
\mmf(\C) := \overline{\Q}^T\R\overline{\Q} ~\quad\text{with}\quad~ \overline{\Q} = \Q^L\ldots\Q^2\Q^1,
\end{equation}
where $\Q^\ell \in \mathcal{O}$ and $\Q^\ell_{[m]\setminus \Sind_{\ell-1},[m]\setminus \Sind_{\ell-1}}\!\! = \I_{m-d_\ell}$ 
for some nested sequence of sets $[m] = \Sind_0 \supseteq \Sind_1 \supseteq \ldots \supseteq \Sind_L$ 
with $|\Sind_\ell| = d_\ell$ and $\R \in \mathcal{R}^m_{\Sind_L}$. 
\end{definition}


\noindent $\Sind_{\ell-1}$ is referred to as the `active set' at the $\ell^{th}$ level, since $\Q^\ell$ is identity outside $[m]\setminus \Sind_{\ell-1}$.
The nesting of the $\Sind_\ell$s implies that after applying $\Q^\ell$ at some level $\ell$, 
$\Sind_{\ell-1}\setminus \Sind_{\ell}$ rows/columns are removed from the active set, and are not operated on subsequently.
This active set trimming is done at all $L$ levels, 
leading to a nested subspace interpretation for the sequence of compressions $\C^\ell = \Q^\ell\C^{\ell-1}(\Q^\ell)^T$ ($\C^0 = \C$ and $\R = \C^L$).
In fact, \cite{kondor2014} has shown that, for a general class of symmetric matrices, 
MMF from Definition \ref{thm:mmfdef} entails a Mallat style multiresolution analysis (MRA) \cite{mallat}.
Observe that depending on the choice of $\Q^\ell$, only a few dimensions of $\C^{\ell-1}$ are forced to interact, 
and so the composition of rotations is hypothesized to extract subtle or softer notions of structure in $\C$.


Since multiresolution is represented as matrix factorization here (see \eqref{eq:mmfdef}), the $\Sind_{\ell-1}\setminus\Sind_\ell$ columns of $\overline{\Q}$ correspond to ``wavelets''.
While $d_1,d_2,\ldots$ can be any monotonically decreasing sequence, 
we restrict ourselves to the simplest case of $d_\ell = m-\ell$. 
Within this setting, the number of levels $L$ is at most $m-k+1$, and each level contributes a single wavelet. 
Given $\Sind_1,\Sind_2,\ldots$ and $\mathcal{O}$, the matrix factorization of \eqref{eq:mmfdef} reduces to determining the $\Q^\ell$ rotations and the residual $\R$, which is usually done by 
minimizing the squared Frobenius norm error   
\begin{equation}\label{eq:mmfobj}
\min_{\tiny \Q^\ell \in \mathcal{O}, \R \in \mathcal{R}^m_{S_L}} \quad \| \C - \mmf(\C)\|^2_{\textrm{Frob}}.  
\end{equation}
The above objective can be decomposed as a sum of contributions from each of the $L$ different levels (see Proposition $1$, \cite{kondor2014}), 
which suggests computing the factorization in a greedy manner as $\C=\C^0\mapsto \C^1\mapsto \C^2\mapsto \ldots \mapsto \R$. 
This error decomposition is what drives much of the intuition behind our algorithms. 


After $\ell-1$ levels, $\C^{\ell-1}$ is the compression and $\Sind_{\ell-1}$ is the active set. 
In the simplest case of $\mathcal{O}$ being the class of so-called $k$--point rotations 
(rotations which affect at most $k$ coordinates) and $d_\ell = m-\ell$, 
at level $\ell$ the algorithm needs to determine three things: 
(a) the $k$--tuple $t^\ell$ of rows/columns involved in the rotation, 
(b) the nontrivial part $\Orth:=\Q^\ell_{t^\ell,t^\ell}$ of the rotation matrix, and 
(c) $s^\ell$, the index of the row/column that is subsequently designated a wavelet and removed from the active set.  
%
Without loss of generality, let $s^\ell$ be the last element of $t^\ell$. 
Then the contribution of level $\ell$ to the squared Frobenius norm error (\ref{eq:mmfobj}) is (see supplement)  
\begin{equation}\begin{aligned}\label{eq:mmferr}
&\err(\C^{\ell-1}; \Orth^\ell; t^\ell, s) = 
2\sum_{i=1}^{k-1} [\Orth\C^{\ell-1}_{t^\ell,t^\ell}\Orth^T]_{k,i}^2 \\
& + 2[\Orth\B\B^T\Orth^T]_{k,k} \quad\text{where}\quad \B = \C^{\ell-1}_{t^\ell,\Sind_{\ell-1}\setminus t^\ell}, 
\end{aligned}\end{equation}
and, in the definition of $\B$, $t^\ell$ is treated as a set. 
The factorization then works 
by minimizing this quantity in a greedy fashion, i.e., 
\begin{equation} \begin{aligned}\label{eq:greedy}
& \Q^\ell, t^\ell, s^\ell \leftarrow \argmin_{\Orth, t, s} \err(\C^{\ell-1}; \Orth; t, s)\\
& \Sind_\ell \leftarrow \Sind_{\ell-1}\setminus s^\ell \quad;\quad \C^\ell = \Q^\ell\C^{\ell-1}(\Q^\ell)^T. 
\end{aligned} \end{equation}


\section{Incremental MMF} \label{sec:online}

We now motivate our algorithm using \eqref{eq:mmferr} and \eqref{eq:greedy}.
Solving \eqref{eq:mmfobj} amounts to estimating the $L$ different $k$-tuples $t^1,\ldots,t^L$ sequentially.
At each level, the selection of the best $k$-tuple is clearly combinatorial, making the exact MMF computation 
(i.e., explicitly minimizing \eqref{eq:mmfobj}) very costly even for $k =3$ or $4$ (this has been independently observed in \cite{mmc}).
As discussed in Section \ref{sec:intro}, higher order MMFs (with large $k$) are nevertheless inevitable for allowing arbitrary interactions among dimensions 
(see supplement for a detailed study), 
and our proposed incremental procedure exploits some interesting properties of the factorization error and other redundancies in $k$-tuple computation. 
The core of our proposal is the following setup. 


\subsection{Overview} \label{sec:overview}

\noindent Let $\tC\in\mathbb{R}^{(m+1)\times (m+1)}$ be the extension of $\C$ by a {\it single} new column $\wvec=[\uvec^T\!,v]^T$, which manipulates $\C$ as: 
\begin{equation}\label{eq:c-ctilde} 
\tC=
\left[
\begin{array}{c|c}
 \C & \uvec \\
\hline
\uvec^T & v
\end{array}
\right]\:.
\end{equation}
The goal is to compute $\mmf(\tC)$. 
Since $\C$ and $\tC$ share all but one row/column (see \eqref{eq:c-ctilde}), if we have access to $\mmf(\C)$, one should, in principle, 
be able to modify $\C$'s underlying sequence of rotations to construct $\mmf(\tC)$. 
This {\it avoids} having to recompute everything for $\tC$ from scratch, i.e., 
performing the greedy decompositions from \eqref{eq:greedy} on the entire $\tC$. 


The hypothesis for manipulating $\mmf(\C)$ to compute $\mmf(\tC)$ comes from the precise computations involved in the factorization.
Recall \eqref{eq:mmferr} and the discussion leading up to the expression. 
At level $\ell+1$, the factorization picks the `best' candidate rows/columns from $\C^\ell$ that correlate the most with each other, 
so that the resulting diagonalization induces the smallest possible off-diagonal error over the rest of the active set.
The components contributing towards this error are driven by the inner products $(\C_{:,i}^\ell)^T\C_{:,j}^\ell$ for some columns $i$ and $j$.
In some sense, the largest such correlated rows/columns get picked up, and adding one new entry to $\C^\ell_{:,i}$ may not change the {\it range} of these correlations.
Extending this intuition across all levels, we argue that
\begin{equation}\label{eq:hypo} \argmax_{i,j} \tC_{:,i}^T\tC_{:,j} \approx \argmax_{i,j} \C_{:,i}^T\C_{:,j}. 	 
\end{equation}
Hence, the $k$-tuples computed from $\C$'s factorization are reasonably good candidates even after introducing $\wvec$. 
To better formalize this idea, and in the process present our algorithm, we parameterize the output structure of $\mmf(\C)$ in terms of the sequence of rotations and the wavelets. 


\subsection{The graph structure of \mathversion{bold}$\mmf(\C)$\mathversion{normal}} \label{sec:mmfgraph}

\begin{figure}\centering 
\includegraphics[width=85mm]{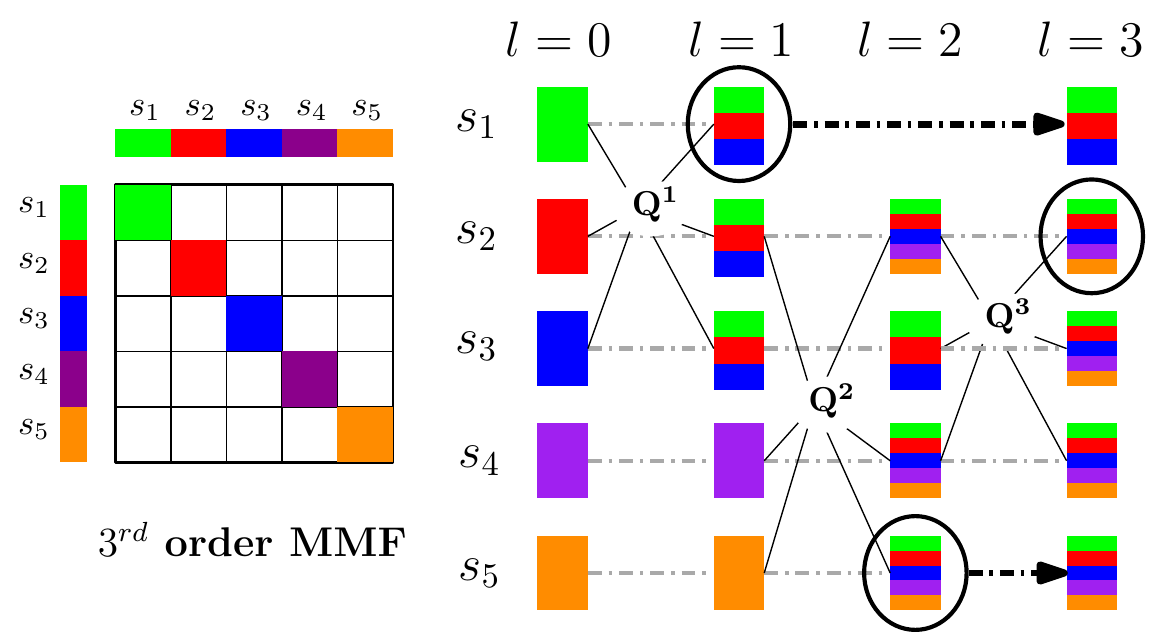} 
\caption{\footnotesize \label{fig:example} An example $5\times 5$ matrix, and its $3^{rd}$ order MMF graph (better in color).
  $\Q^1$, $\Q^2$ and $\Q^3$ are the rotations. $s_1$, $s_5$ and $s_2$ are wavelets (marked as black ellipses) at $l=1$, $2$ and $3$ respectively.
  The arrows imply that a wavelet is not involved in future rotations.}
\end{figure}

If one has access to the sequence of $k$-tuples $t^1,\ldots,t^L$ involved in the rotations and the 
corresponding wavelet indices ($s^1,\ldots,s^L$), 
then the factorization is straightforward to compute i.e., there is no greedy search anymore. 
Recall that by definition $s^\ell \in t^\ell$ and $s^\ell \notin \Sind_\ell$ (see \eqref{eq:greedy}). 
To that end, for a given $\mathcal{O}$ and $L$, $\mmf(\C)$ can be `equivalently' represented using a depth $L$ {\it MMF graph} $\graph(\C)$. 
Each level of this graph shows the $k$-tuple $t^\ell$ involved in the rotation, and the corresponding wavelet $s^\ell$ i.e., $\graph(\C) := \{t^\ell, s^\ell\}_1^L$.
Interpreting the factorization in this way is notationally convenient for presenting the algorithm.
More importantly, such an interpretation is central for visualizing hierarchical dependencies among dimensions of $\C$, and will be discussed in detail in Section \ref{sec:graphinterp}.
An example of such a $3^{rd}$ order MMF graph constructed from a $5 \times 5$ matrix is shown in Figure \ref{fig:example} (the rows/columns are color coded for better visualization).
At level $\ell=1$, $s_1$, $s_2$ and $s_3$ are diagonalized while designating the rotated $s_1$ as the wavelet. 
This process repeats for $\ell=2$ and $3$. 
As shown by the color-coding of different compositions, 
MMF gradually teases out {\it higher-order} correlations that can only be revealed after composing the rows/columns at one or more scales (levels here).


For notational convenience, we denote the MMF graphs of $\C$ and $\tC$ as $\graph := \{t^\ell, s^\ell\}_1^L$ and $\tgraph := \{\ttt^\ell, \ts^\ell\}_1^{L+1}$. 
Recall that $\tgraph$ will have one more level than $\graph$ since the row/column $\wvec$, indexed $m+1$ in $\tC$, is being added (see \eqref{eq:c-ctilde}).
The goal is to estimate $\tgraph$ without recomputing all the $k$-tuples using the greedy procedure from \eqref{eq:greedy}. 
This translates to {\it inserting} the new index $m+1$ into the $t^\ell$s and modifying $s^\ell$s accordingly.
Following the discussion from Section \ref{sec:overview}, incremental MMF argues that inserting this one new element into the graph will not result in global changes in its topology. 
Clearly, in the pathological case, $\graph$ may change arbitrarily, 
but as argued earlier (see discussion about \eqref{eq:hypo}) the chance of this happening for non-random matrices with reasonably large $k$ is small. 
The core operation then is to compare the {\it new} $k$-tuples resulting from the addition of $\wvec$ to the best ones from $[m]^k$ provided via $\graph$.
If the newer $k$-tuple gives better error (see \eqref{eq:mmferr}), then it will {\it knock out} an existing $k$-tuple.
This constructive insertion and knock-out procedure is the incremental MMF.


\subsection{Inserting a new row/column} \label{sec:insone}

The basis for this incremental procedure is that one has access to $\graph$ (i.e., MMF on $\C$). 
We first present the algorithm assuming that this ``initialization'' is provided, and revisit this aspect shortly. 
The procedure starts by setting $\ttt^\ell = t^\ell$ and $\ts^\ell = s^\ell$ for $\ell=1,\ldots,L$.
Let $\ins$ be the set of elements (indices) that needs to be inserted into $\graph$.
At the start (the first level) $\ins = \{m+1\}$ corresponding to $\wvec$. 
Let $\ttt^1 = \{p_1,\ldots,p_k\}$. The new $k$-tuples that account for inserting entries of $\ins$ are 
$\{m+1\} \cup t^1 \setminus p_i$ ($i=1,\ldots,k$).
These new $k$ candidates are the probable alternatives for the existing $\ttt^1$.
Once the best among these $k+1$ candidates is chosen, an existing $p_i$ from $\ttt^1$ may be knocked out.


If $\ts^1$ gets knocked out, then $\ins = \{\ts^1\}$ for future levels. 
This follows from MMF construction, where wavelets at $\ell^{th}$ level are not involved in later levels. 
Since $\ts^1$ is knocked out, it is the new inserting element according to $\graph$.
On the other hand, if one of the $k-1$ scaling functions is knocked out, $\ins$ is not updated. 
This simple process is repeated sequentially from $\ell=1$ to $L$.
At $L+1$, there are no estimates for $\ttt^{L+1}$ and $\ts^{L+1}$, and so, the procedure simply selects the best $k$-tuple from the remaining active set $\tSind_L$.
Algorithm \ref{alg:olmmf} summarizes this insertion and knock-out procedure.
\begin{algorithm}
\caption{\label{alg:olmmf}~~\textsc{InsertRow}$(\C, \wvec, \{ t^\ell, s^\ell \}_{\ell=1}^L)$} 
\begin{algorithmic} \label{alg:olmmf}
\ENSURE $\{ \ttt^\ell, \ts^\ell \}_{\ell=1}^{L+1}$ 
\STATE $\tC^0 \leftarrow \tC$~ as in \eqref{eq:c-ctilde}
\STATE  $z^1 \leftarrow m+1$ 
\FOR {$\ell=1$~~to~~$L-1$}
\STATE $\{\ttt^\ell\!,\ts^\ell,z^{\ell+1}\!,\Q^\ell\}$  
$\!\leftarrow\!\!\textsc{CheckInsert}(\tC^{\ell-1}\!;t^\ell\!, s^\ell, z^\ell)$\hspace{-20pt}
\STATE $\tC^\ell = \Q^\ell\tC^{\ell-1}(\Q^\ell)^T$
\ENDFOR
\STATE $\mathcal{T} \leftarrow \textsc{GenerateTuples}([m+1]\setminus \cup_{\ell=1}^{L-1} \ts^\ell(\tC))$ 
\STATE $\{\tilde{\Orth},\ttt^L,\ts^L\} \leftarrow \argmin_{\Orth, t\in\mathcal{T}, s\in t} \err(\tC^{L-1};\Orth;t,s)$\hspace{-20pt}
\STATE $\Q^L = \I_{m+1}$,~~ $\Q^L_{\ttt^L,\ttt^L} = \tilde{\Orth}$,~~ $\tC^{L} = \Q^L\tC^{L-1}(\Q^L)^T$
\end{algorithmic} 
\end{algorithm}
\vspace{-3.5mm}
%
\begin{algorithm}
\caption{\label{alg:insert} ~~\textsc{CheckInsert}$(\A, \htt, \hs, z)$}
\begin{algorithmic} \label{alg:insert}
\ENSURE $\ttt$, $\ts$, $z$, $\Q$
\STATE $\mathcal{T} \leftarrow \textsc{GenerateTuples}(\htt,z)$ 
\STATE $\{\tilde{\Orth},\ttt,\ts\} \leftarrow \argmin_{\Orth, t\in\mathcal{T}, s\in t} \err(\A;\Orth;t,s)$
\IF {$\ts \in z$} 
\STATE $z \leftarrow (z \cup \hs) \setminus \ts$ 
\ENDIF
\STATE $\Q = \I_{m+1}$,~~$\Q_{\ttt,\ttt} = \tilde{\Orth}$
\end{algorithmic}
\end{algorithm}


\subsection{Incremental MMF Algorithm} \label{sec:insall}

Observe that Algorithm \ref{alg:olmmf} is for the setting from \eqref{eq:c-ctilde} where one extra row/column is added to a given MMF, 
and clearly, the incremental procedure can be repeated as more and more rows/columns are added.
Algorithm \ref{alg:multiolmmf} summarizes this incremental factorization for arbitrarily large and dense matrices.
It has two components: an initialization on some randomly chosen small block (of size $\tm \times \tm$) of the entire matrix $\C$; 
followed by insertion of the remaining $m-\tm$ rows/columns using Algorithm \ref{alg:olmmf} in a streaming fashion 
(similar to $\wvec$ from \eqref{eq:c-ctilde}).
The initialization entails computing a batch-wise MMF on this small block ($\tm \geq k$).

\textsc{BatchMMF:}
Note that at each level $\ell$, the error criterion in \eqref{eq:mmferr} can be explicitly minimized via an exhaustive search 
over all possible $k$-tuples from $\Sind_{\ell-1}$ (the active set) and a randomly chosen (using properties of $QR$ decomposition \cite{mezzadri2006generate}) dictionary of $k^{th}$ order rotations.
If the dictionary is large enough, the exhaustive procedure would lead to the smallest possible decomposition error (see \eqref{eq:mmfobj}).
However, it is easy to see that this is combinatorially large, with an overall complexity of $\mathcal{O}(n^k)$ \cite{pmmf} and will not scale well beyond $k = 4$ or so. 
Note from Algorithm \ref{alg:olmmf} that the error criterion $\err(\cdot)$ in this second stage which inserts the rest of the $m-\tm$ rows is performing an exhaustive search as well.
\begin{algorithm}
\caption{\label{alg:multiolmmf}~~ \textsc{Incremental MMF}(\C)}
\begin{algorithmic} 
\ENSURE $\mmf(\C)$
\STATE $\bar{\C} = \C_{[\tm],[\tm]}$, $L = m-k+1$
\STATE $\{ t^\ell, s^\ell \}_1^{\tm-k+1} \leftarrow \textsc{BatchMMF}(\bar{\C})$
\FOR {$j \in \{\tm+1,\ldots,m\}$} 
\STATE $\{ t^\ell, s^\ell \}_1^{j-k+1}$
$\leftarrow \textsc{InsertRow}(\bar{\C}, \C_{j,:}, \{ t^\ell, s^\ell \}_1^{j-k})$ 
\STATE $\bar{\C} = \C_{[j],[j]}$ 
\ENDFOR
\STATE $\mmf(\C) := \{ t^\ell, s^\ell \}_1^L$
\end{algorithmic}
\end{algorithm}


\noindent {\bf Other Variants:}
The are two alternatives that avoid this exhaustive search. Since $\Q^\ell$'s job is to diagonalize some $k$ rows/columns (see Definition \ref{thm:mmfdef}),
one can simply pick the relevant $k \times k$ block of $\C^\ell$ and compute the best $\Orth$ (for a given $t^\ell$).
Hence the first alternative is to bypass the search over $\mathcal{O}$ (in \eqref{eq:greedy}), and simply use the eigen-vectors of $\C^\ell_{t^\ell,t^\ell}$ for some tuple $t^\ell$. 
Nevertheless, the search over $\Sind_{\ell-1}$ for $t^\ell$ still makes this approximation reasonably costly.
Instead, the $k$-tuple selection may be approximated while keeping the exhaustive search over $\mathcal{O}$ intact \cite{pmmf}.
Since diagonalization effectively nullifies correlated dimensions, the best $k$-tuple can be the $k$ rows/columns that are maximally correlated.
This is done by choosing some $s_1 \sim \Sind_{\ell-1}$ (from the current active set), and picking the rest by 
\begin{equation}\label{eq:nonideal}
s_2,\ldots,s_k\leftarrow \argmin_{s_i\sim \Sind_{\ell-1}\setminus s_1} \sum_{i=2}^{k} \frac{(\C^{\ell-1}_{:,s_1})^T\C^{\ell-1}_{:,s_i}}{\|\C^{\ell-1}_{:,s_1}\|\|\C^{\ell-1}_{:,s_i}\|} \quad 
\end{equation}
This second heuristic (which is related to \eqref{eq:hypo} from Section \ref{sec:overview}) has been shown to be robust \cite{pmmf}, 
however, for large $k$ it might miss some $k$-tuples that are vital to the quality of the factorization.
Depending on $\tm$, and the available computational resources at hand, these alternatives can be used instead of the earlier proposed exhaustive procedure for the initialization.
Overall, the incremental procedure scales efficiently for very large matrices, compared to using the batch-wise scheme on the entire matrix.


\section{Experiments} \label{sec:exps}

We study various computer vision and medical imaging scenarios (see supplement for details) to evaluate the quality of incremental MMF factorization and show its utility.
We first provide evidence for factorization's efficacy in selecting the relevant features of interest for regression.
We then show that the resultant MMF graph is a useful tool for visualizing/decoding the learned task-specific representations.

\subsection{Incremental versus Batch MMF} \label{sec:simulations}

The first set of evaluations compares the incremental MMF to the batch version (including the exhaustive search based and the two approximate variants from Section \ref{sec:insall}).
Recall that MMF error is the off-diagonal norm of $\R$, except for the $\Sind_L\times \Sind_L$ block (see \eqref{eq:mmfdef}), 
and the smaller the error is, the closer the factorization is to being exact (see \ref{thm:mmfdef}).
We observed that the incremental MMFs incur approximately the same error as the batch versions, 
while achieving $\gtrsim 20-25$ times speed-up compared to a single-core implementation of the batch MMF.
Specifically, across $6$ different toy examples and $3$ covariance matrices constructed from real data, 
the loss in factorization error is $\lesssim 4\%$ of $\|\C\|_{\textrm{Frob}}$, with no strong dependence on the fraction of the initialization $\tm$ (see Algorithm \ref{alg:multiolmmf}). 
Due to space restrictions, these simulations are included in the supplement.


\subsection{MMF Scores} \label{sec:rellev}

\begin{figure*}[!t] \centering
\subfloat[LevScore Sampling]{\includegraphics[width=35mm]{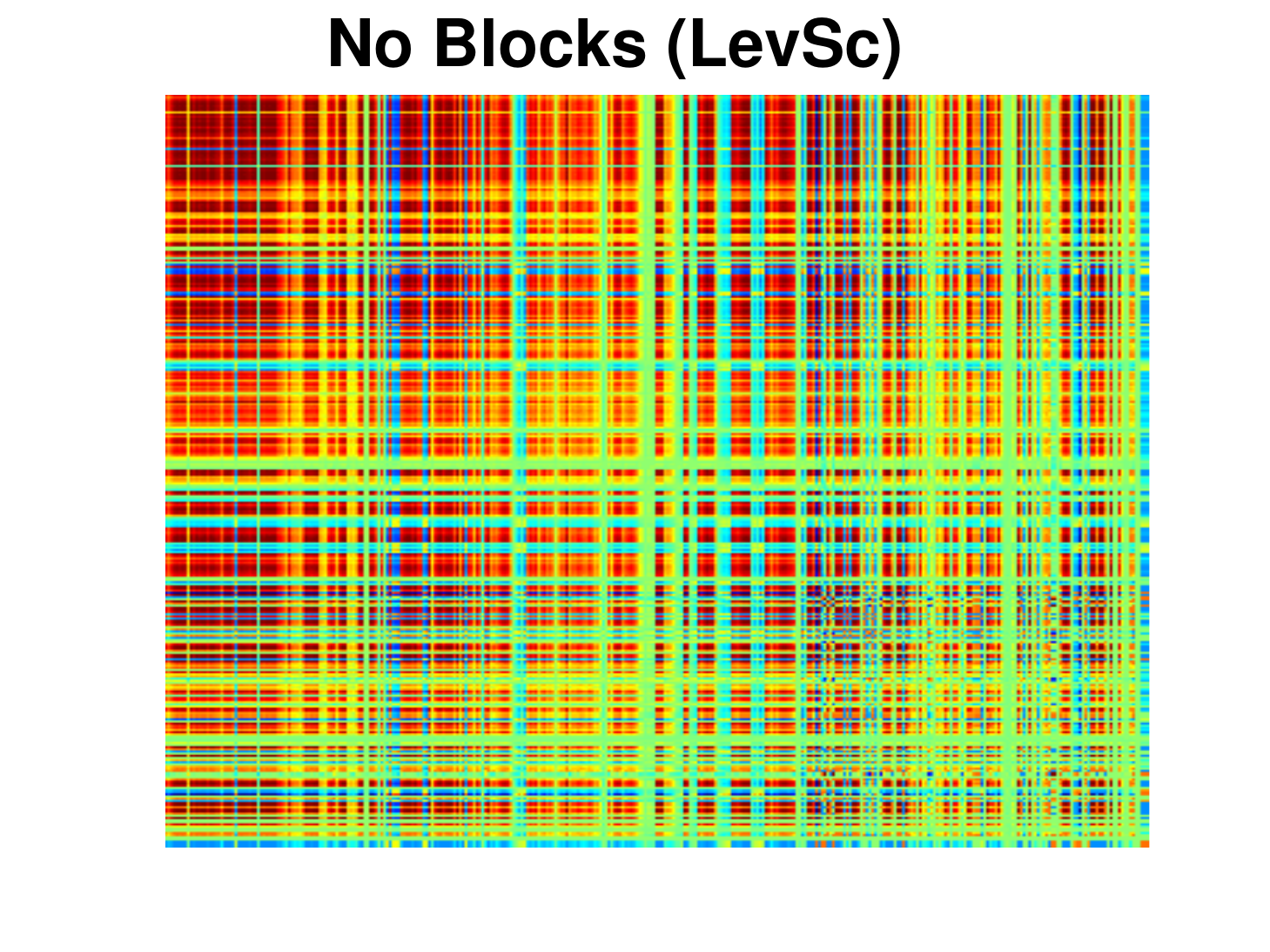}} \hspace{-4mm}
\subfloat[MMFScore Sampling]{\includegraphics[width=35mm]{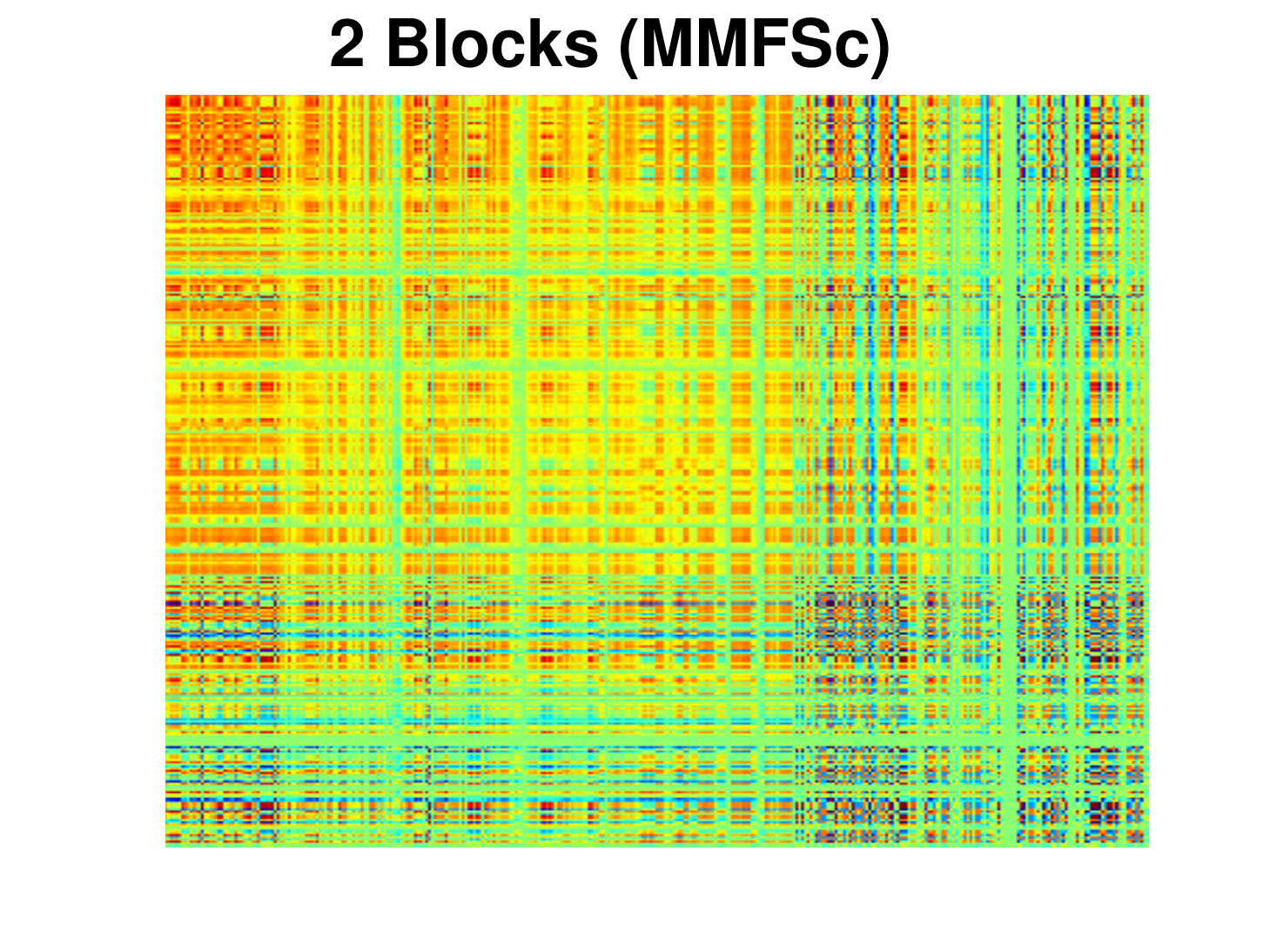}}
\subfloat[Regression Setup -- Constructing the ROI data]{\includegraphics[width=90mm]{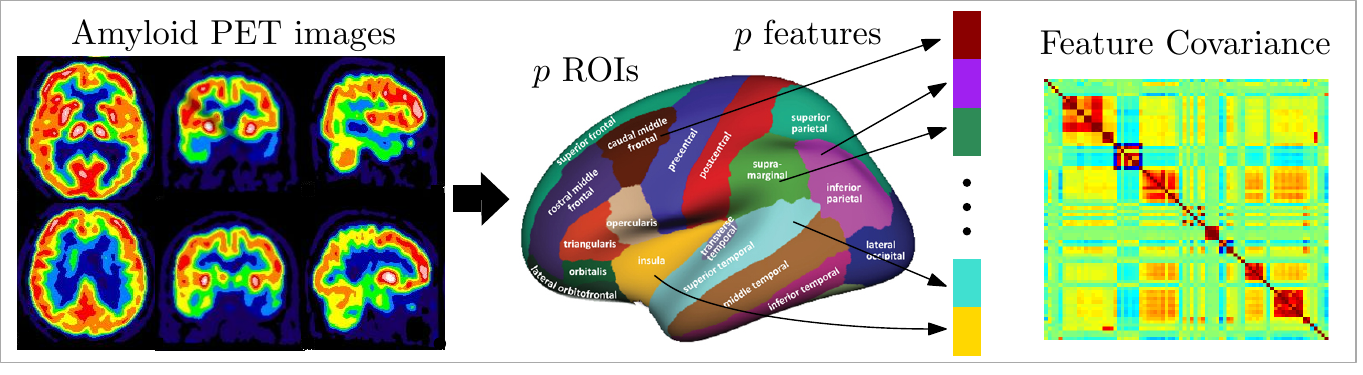}} \\\vspace{-2mm}
\subfloat[$R^2$ vs. $\#$ROIs]{\includegraphics[width=41mm]{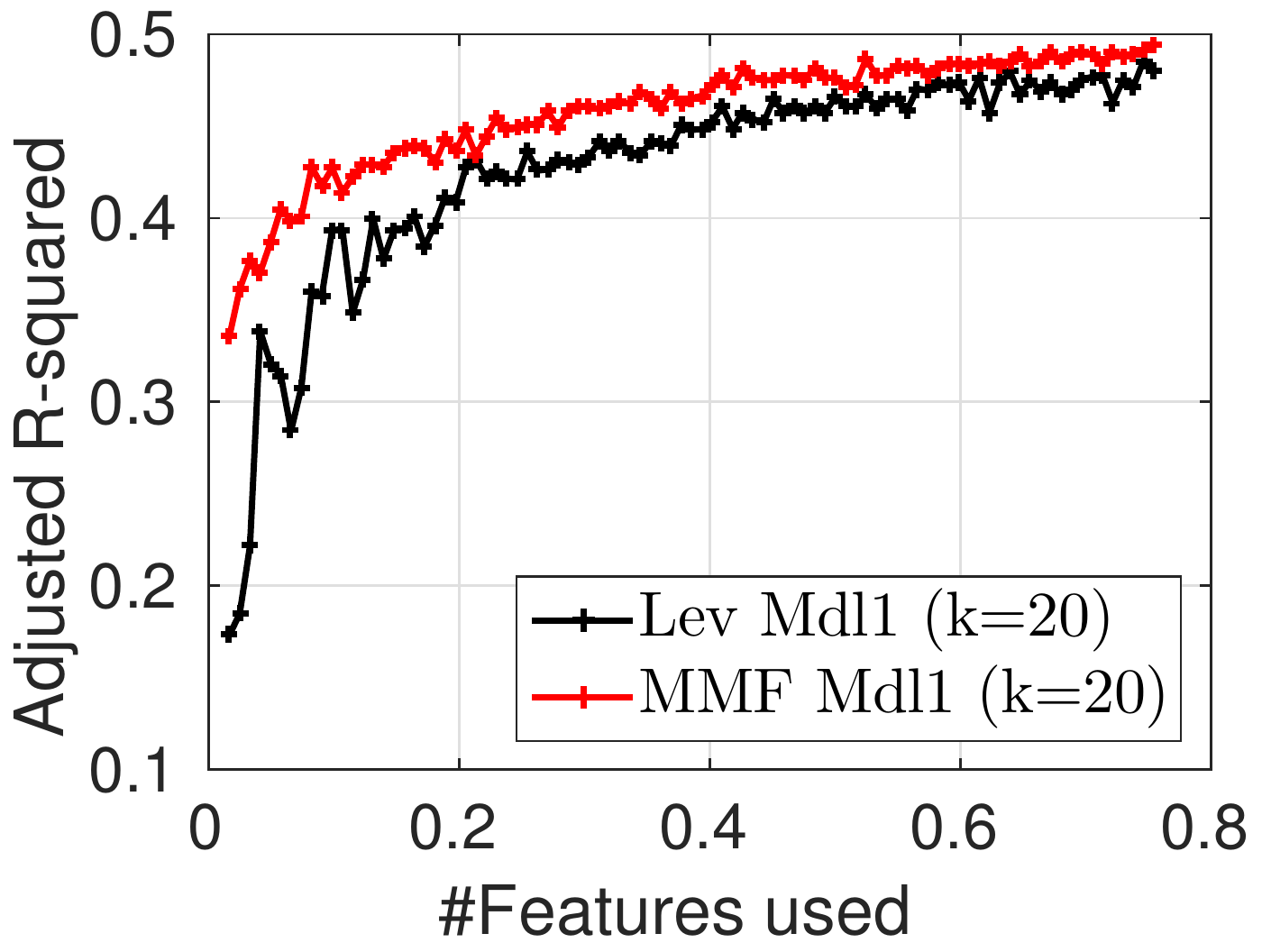}} 
\subfloat[$R^2$ vs. $\#$ROIs]{\includegraphics[width=41mm]{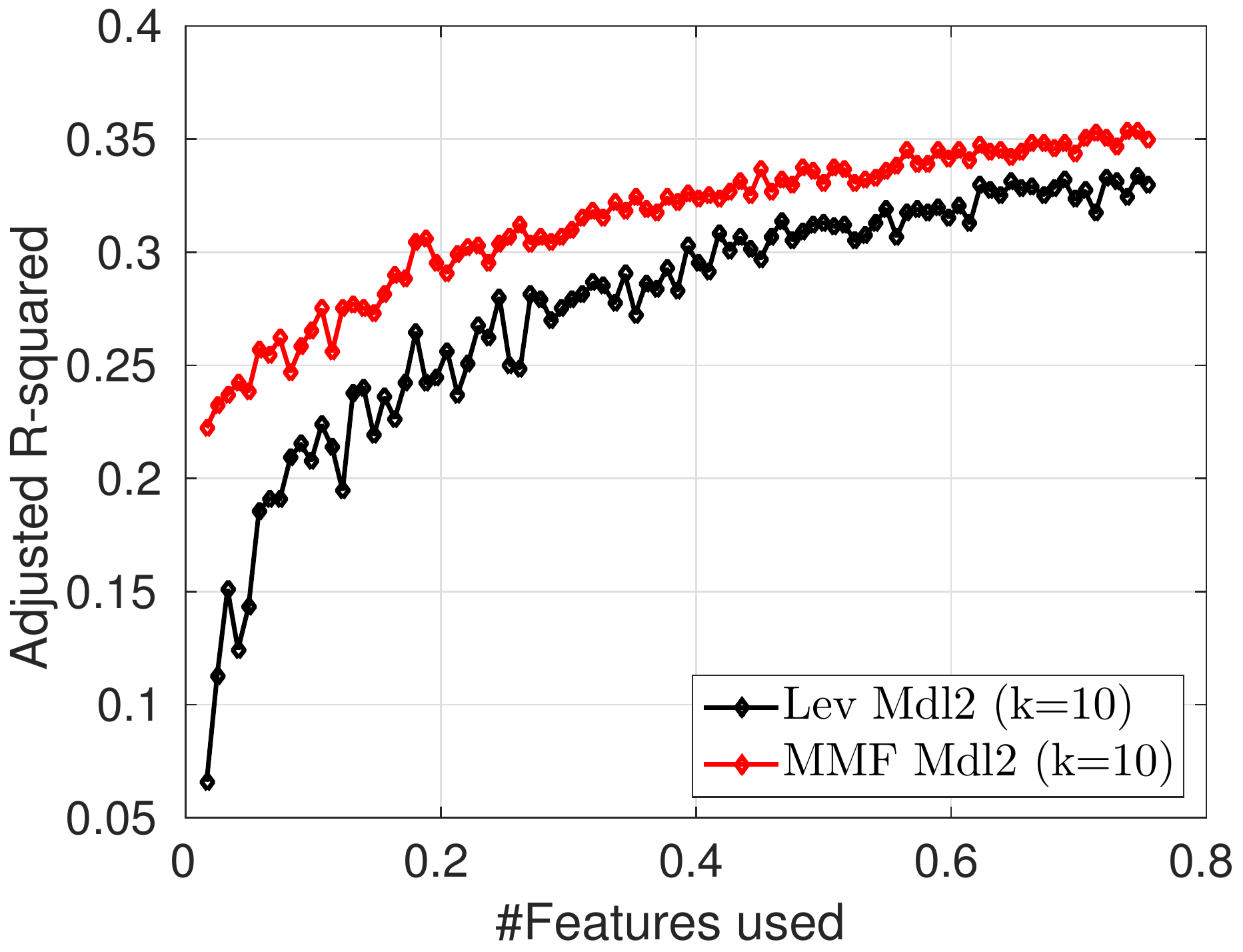}} 
\subfloat[$R^2$ vs. $\#$ROIs]{\includegraphics[width=41mm]{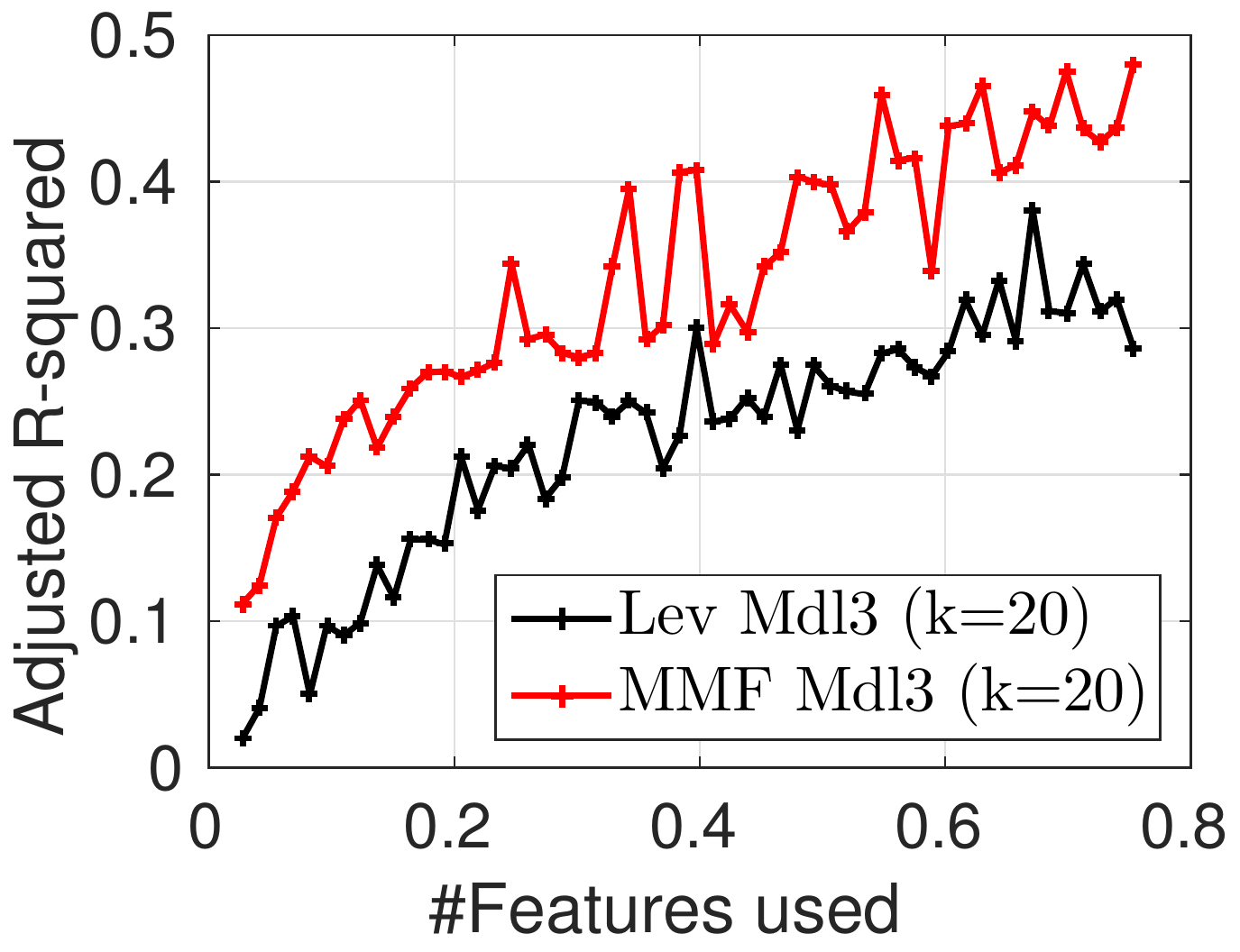}} 
\subfloat[$R^2$ vs. $\#$ROIs]{\includegraphics[width=41mm]{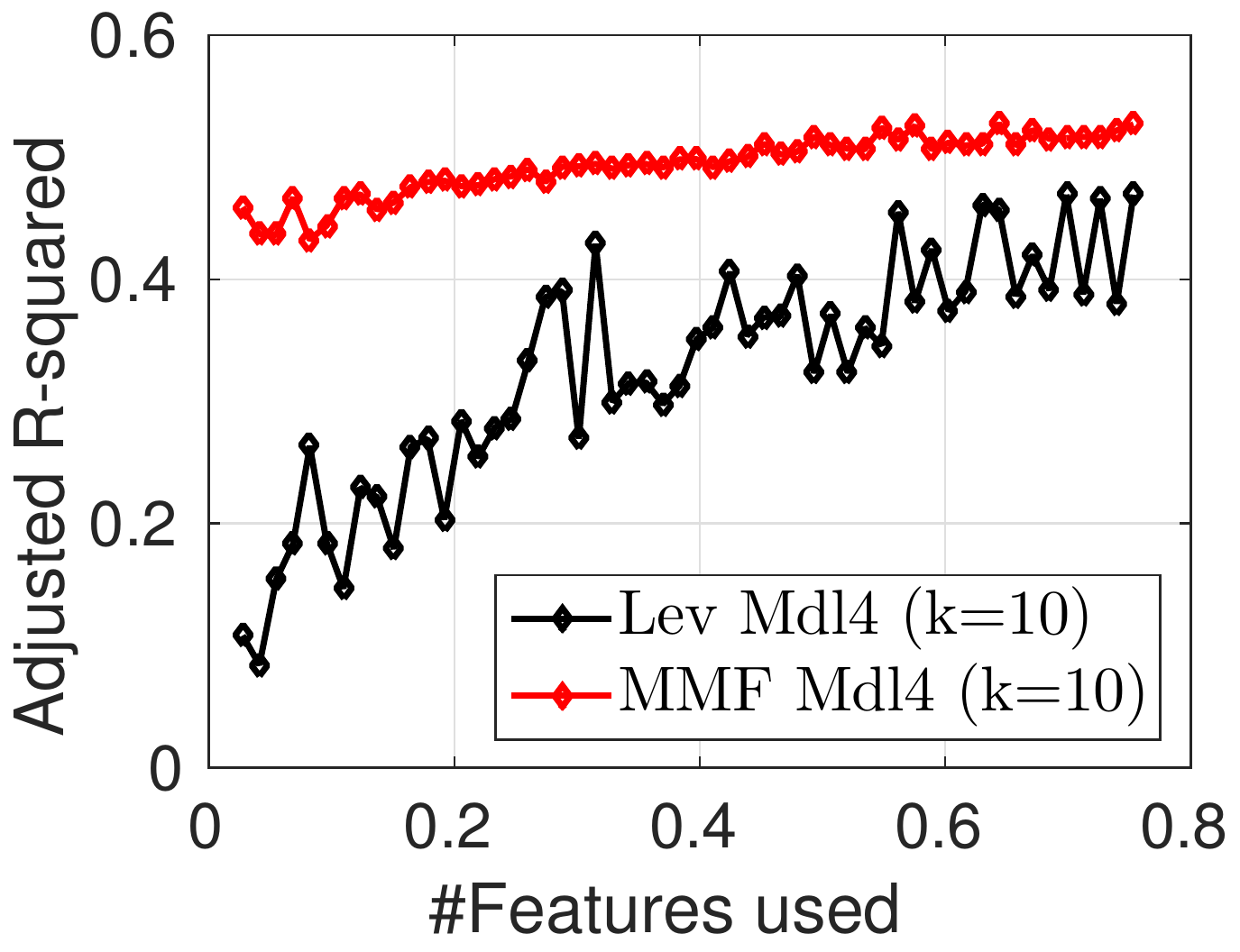}} \\\vspace{-4mm}
\subfloat[$F$ vs. $\#$ROIs]{\includegraphics[width=41mm]{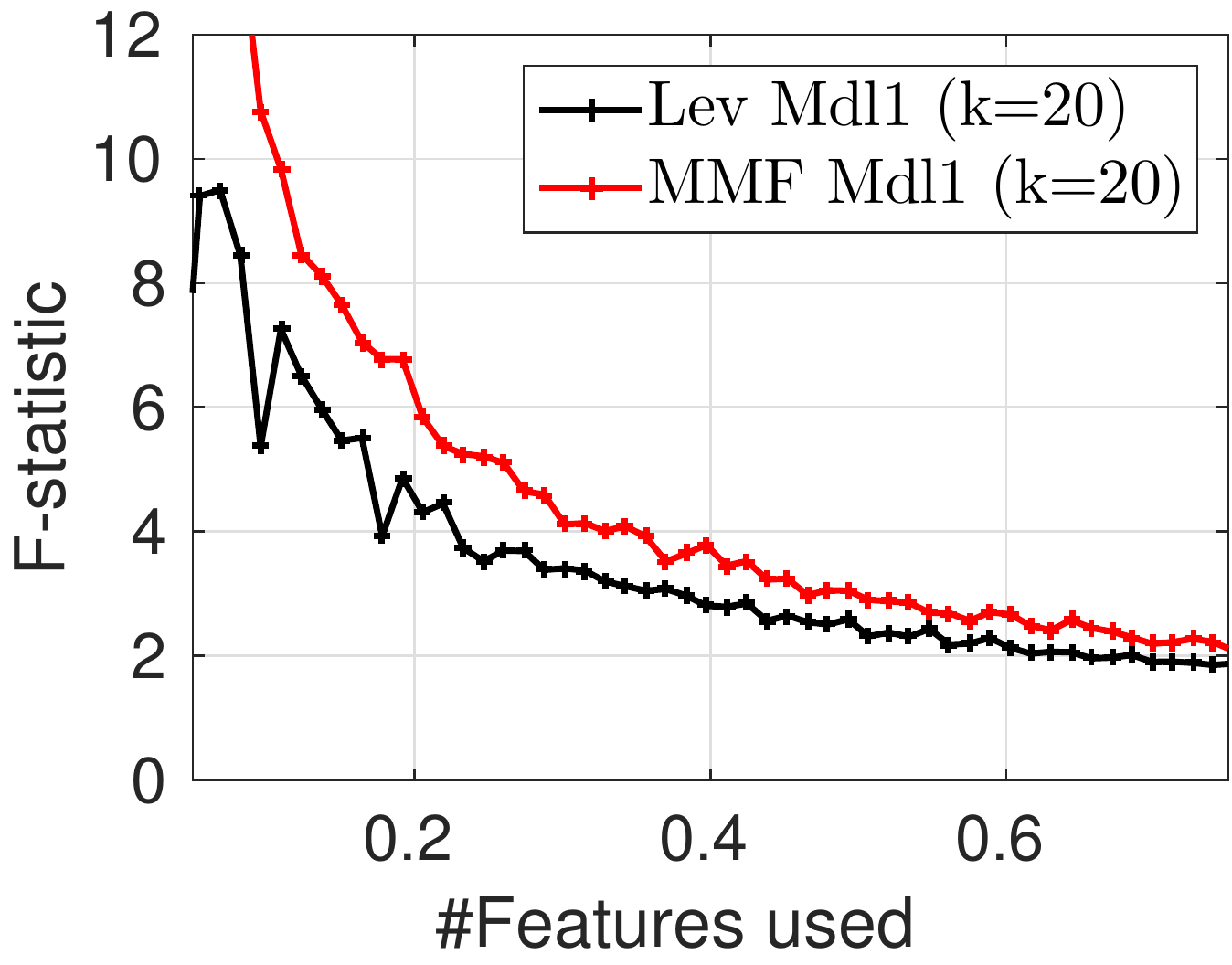}} 
\subfloat[$F$ vs. $\#$ROIs]{\includegraphics[width=41mm]{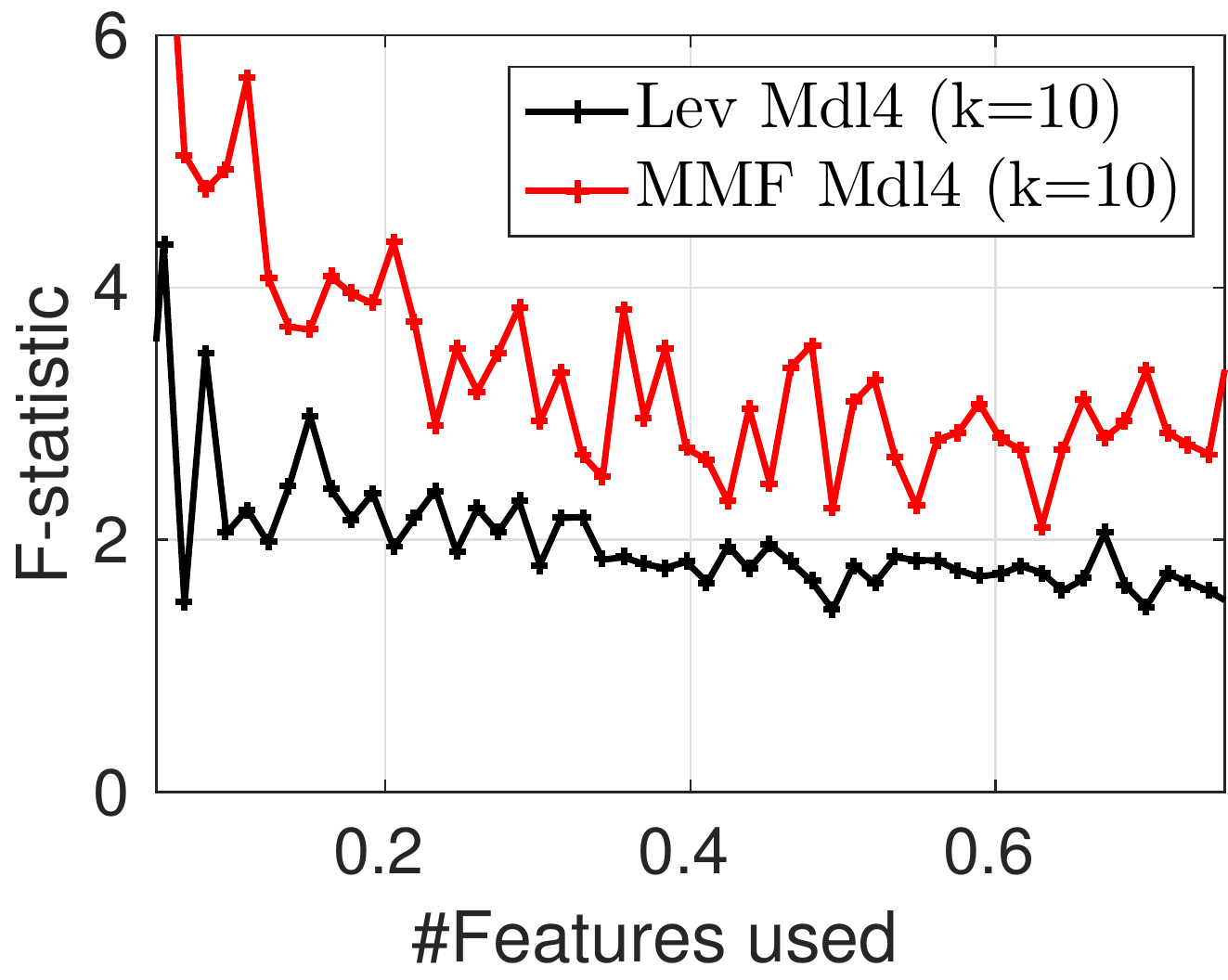}} 
\subfloat[Gain in $R^2$ AUC]{\includegraphics[width=41mm]{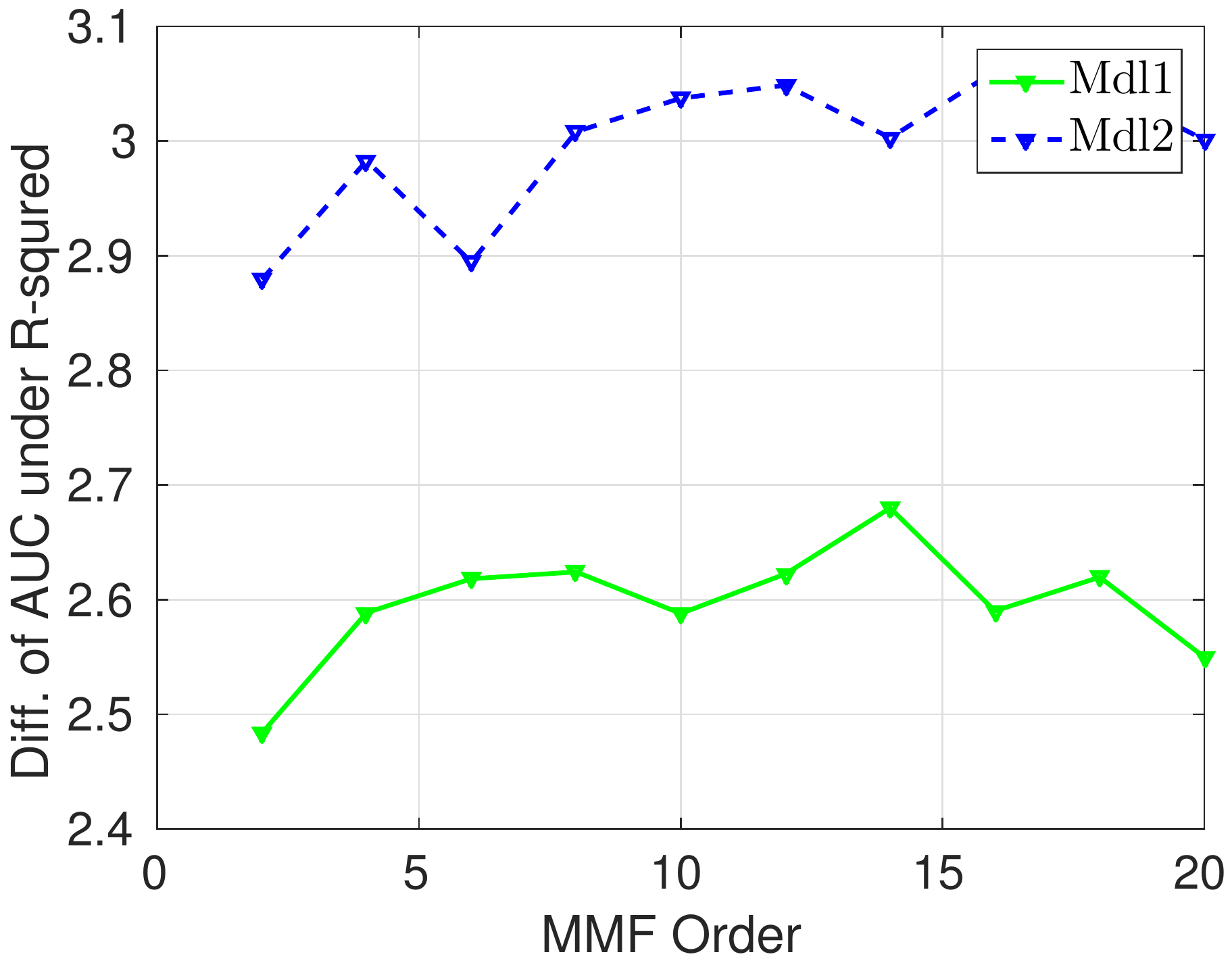}} 
\subfloat[Gain in $R^2$ AUC]{\includegraphics[width=41mm]{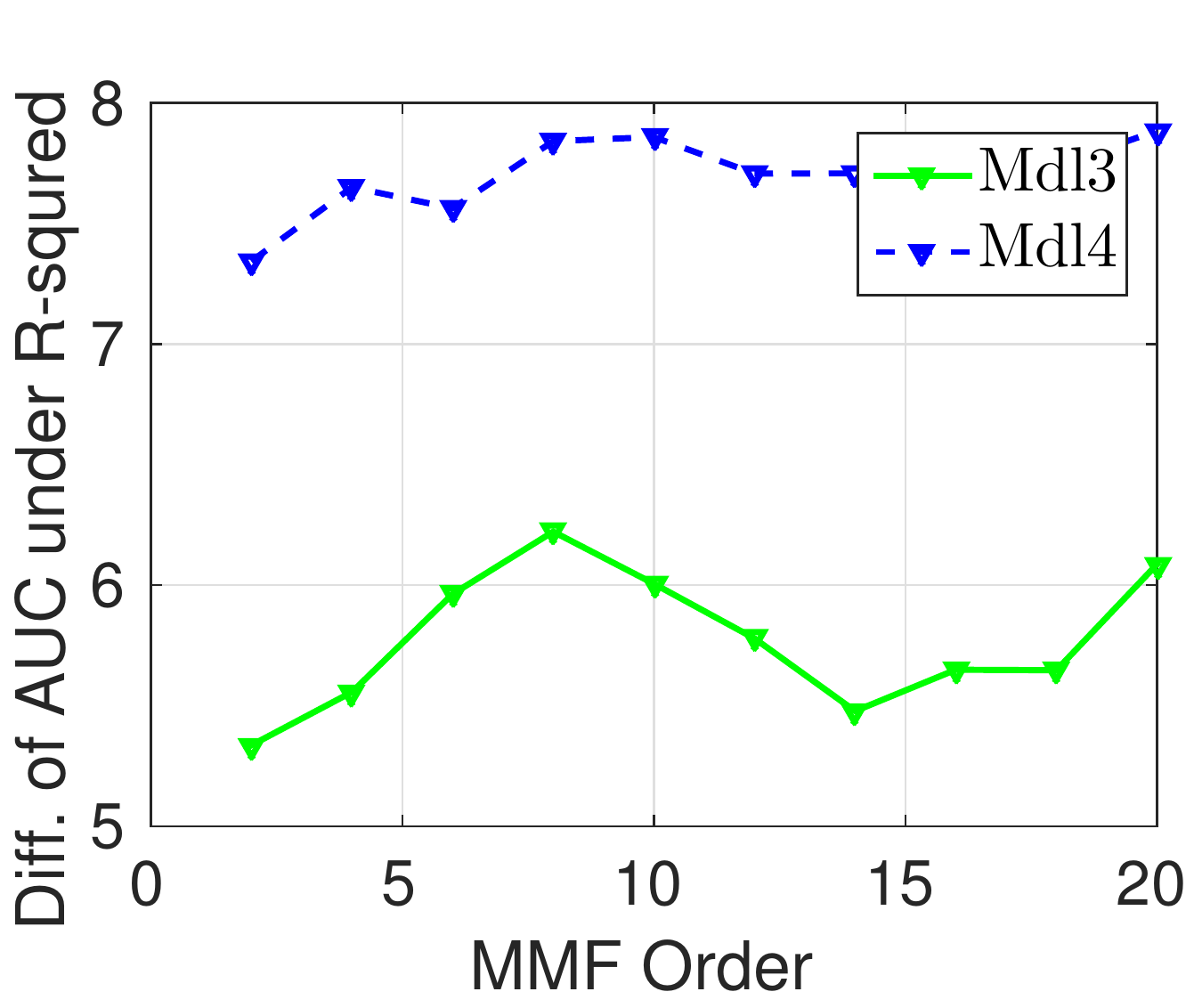}} 
\caption{\footnotesize \label{fig:mmfvslev}
  {\bf Evaluating Feature Importance Sampling of MMF Scores vs. Leverage Scores} (a,b) Visualizing apparent (if any) blocks in instance covariance matrices using best $5\%$ features,
  (c) Regression setup (see structure in covariance), 
  (d-g) Adjusted $R^2$, and (h,i) $F$statistic of linear models, (j,k) gains in $R^2$. Mdl1-Mdl4 are linear models constructed on different datasets (see supplement).
$\tm = 0.1m$ (from Algorithm \ref{alg:multiolmmf}) for these evaluations.}
\end{figure*}

The objective of MMF (see \eqref{eq:mmfobj}) is the signal that is not accounted for by the $k^{th}$-order rotations of MMF (it is $0$ whenever $\C$ is {\it exactly} factorizable).
Hence, $\|(\C-\mmf(\C))_{i,:}\|$ is a measure of the extra information in the $i^{th}$ row that cannot be reproduced by hierarchical compositions of the rest. 
Such value-of-information summaries, referred to as {\it MMF scores}, 
of all the dimensions of $\C$ give an importance sampling distribution, similar to statistical leverage scores \cite{boutsidis2009unsupervised, ma2015statistical}. 
These samplers drive several regression tasks in vision including gesture tracking \cite{rautaray2015vision}, 
face alignment/tracking \cite{cao2014face} and medical imaging \cite{friston1994statistical}. 
Moreover, the authors in \cite{ma2015statistical} have shown that statistical leverage type marginal importance samplers may not be optimal for regression. 
On the other hand, MMF scores give the conditional importance or ``relative leverage'' of each dimension/feature given the remaining ones.
This is because MMF encodes the hierarchical block structure in the covariance, and so, the MMF scores provide better importance samplers than statistical leverages. 
We first demonstrate this on a large dataset with $80$ predictors/features and $1300$ instances.
Figure \ref{fig:mmfvslev}(a,b) shows the instance covariance matrices after selecting the `best' $5\%$ of features. 
The block structure representing the two classes, diseased and non-diseased, is clearly more apparent with MMF score sampling (see the yellow block vs.the rest in Figure \ref{fig:mmfvslev}(b)). 


We exhaustively compared leverages and MMF scores on a medical imaging regression task on region-of-interest (ROI) summaries from positron emission tomography (PET) images. 
The goal is to predict the cognitive score summary using the imaging ROIs.
(see Figure \ref{fig:mmfvslev}(c), and supplement for details).
Despite the fact that the features have high degree of block structure (covariance matrix from Figure \ref{fig:mmfvslev}(c)), 
this information is rarely, if ever, utilized within the downstream models, say multi-linear regression for predicting health status. 
Here we train a linear model using a {\it fraction} of these voxel ROIs sampled according to statistical leverages (from \cite{boutsidis2009unsupervised}) and relative leverage from MMF scores. 
Note that, unlike LASSO, the feature samplers are agnostic to the responses (a setting similar to optimal experimental design \cite{de1995d}). 
The second row of Figure \ref{fig:mmfvslev} shows the Adjusted-$R^2$ of the resulting linear models, and Figure \ref{fig:mmfvslev}(h,i) show the corresponding $F$-statistic.
The $x$-axis corresponds to the fraction of the ROIs selected using the leverage (black lines) and MMF scores (red lines). 
As shown by the red vs. black curves, the voxel ROIs picked by MMF are better both in terms of adjusted-$R^2$ (the explainable variance of the data) and $F$-statistic (the overall significance). 
More importantly, the first few ROIs picked up by MMF scores are more informative than those from leverage scores (left end of $x$-axis in Figure \ref{fig:mmfvslev}(d-i)).
Figure \ref{fig:mmfvslev}(j,k) show the gain in AUC of adjusted-$R^2$ as the order of MMF changes ($x$-axis). Clearly the performance gain of MMF scores is large.
The error bars in these plots are omitted for clarity (see supplement for details, and other plots/comparisons). 
These results show that MMF scores can be used within numerous regression tasks where the number of predictors is large with sample sizes.


\mathversion{bold}
\begin{figure*}[!t] \centering
\subfloat[{\bf $12$ classes}]{\includegraphics[width=43mm]{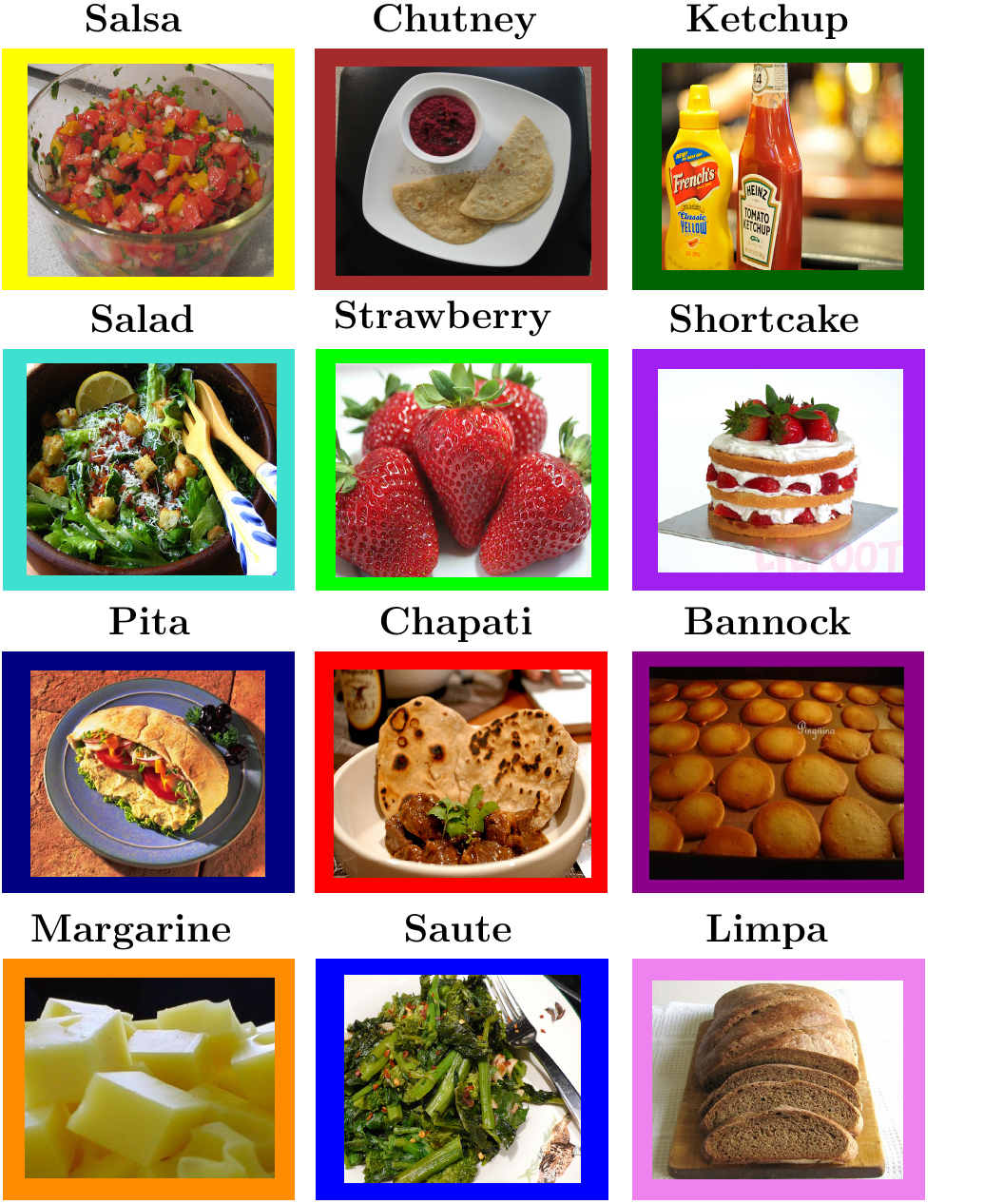}} 
\subfloat[{\bf $5^{th}$ order (FC$7$ layer reps.)}]{\includegraphics[width=63mm]{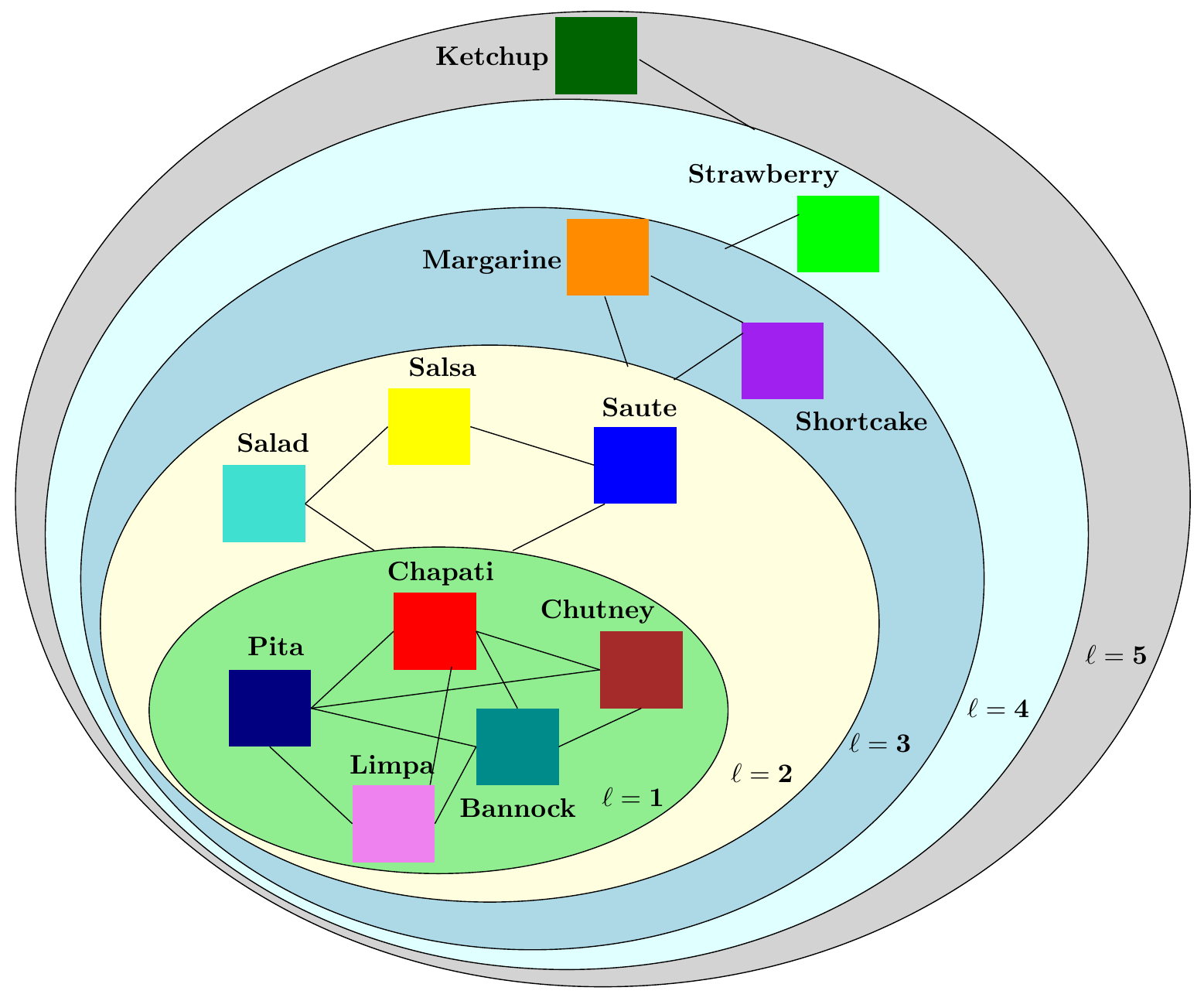}} \hspace{-2mm}
\subfloat[{\bf $5^{th}$ order (FC$7$ layer reps.)}]{\includegraphics[width=63mm]{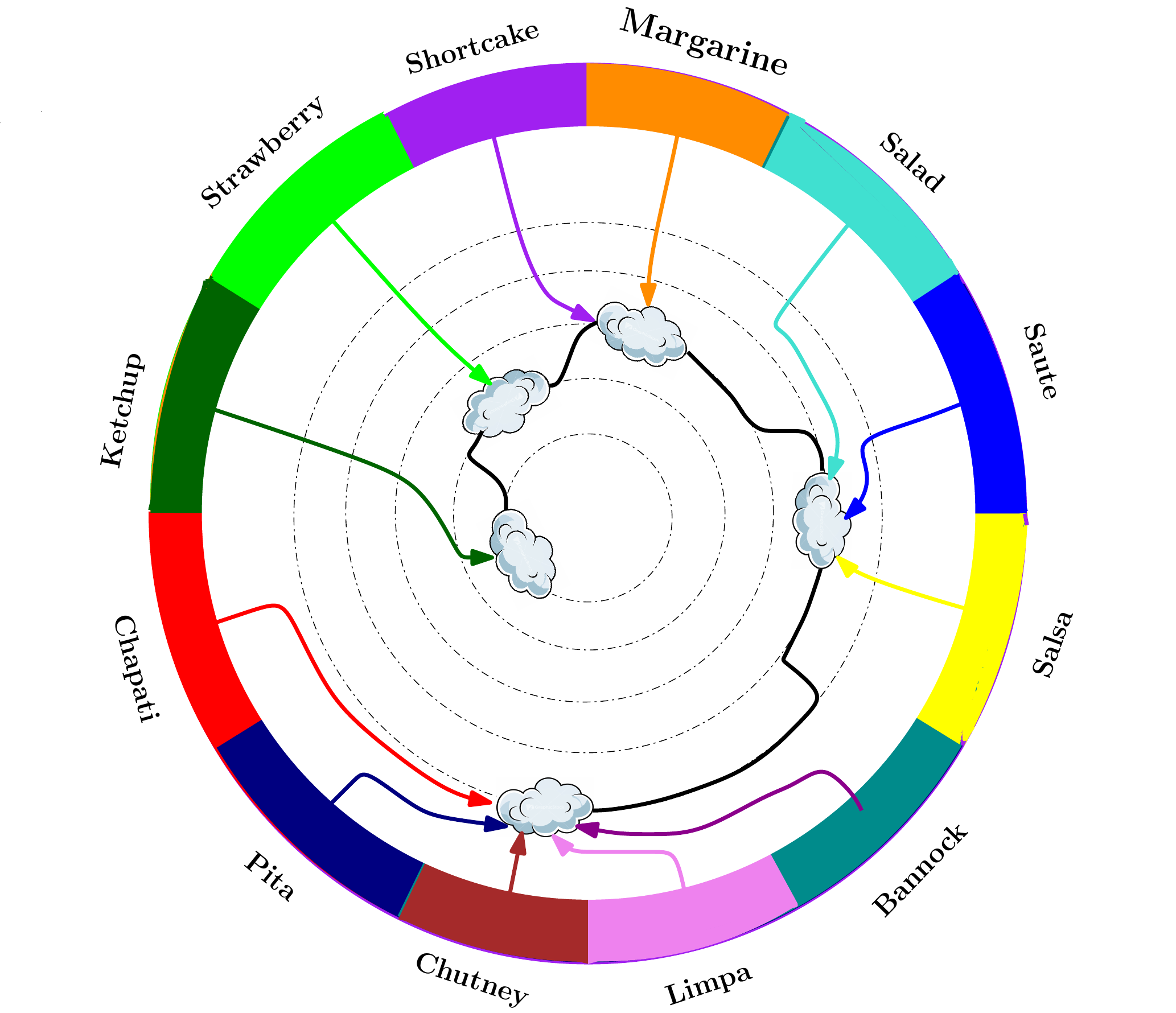}} \\\vspace{-4mm}
\subfloat[{\bf $4^{th}$ order (FC$7$ layer reps.)}]{\includegraphics[width=56mm]{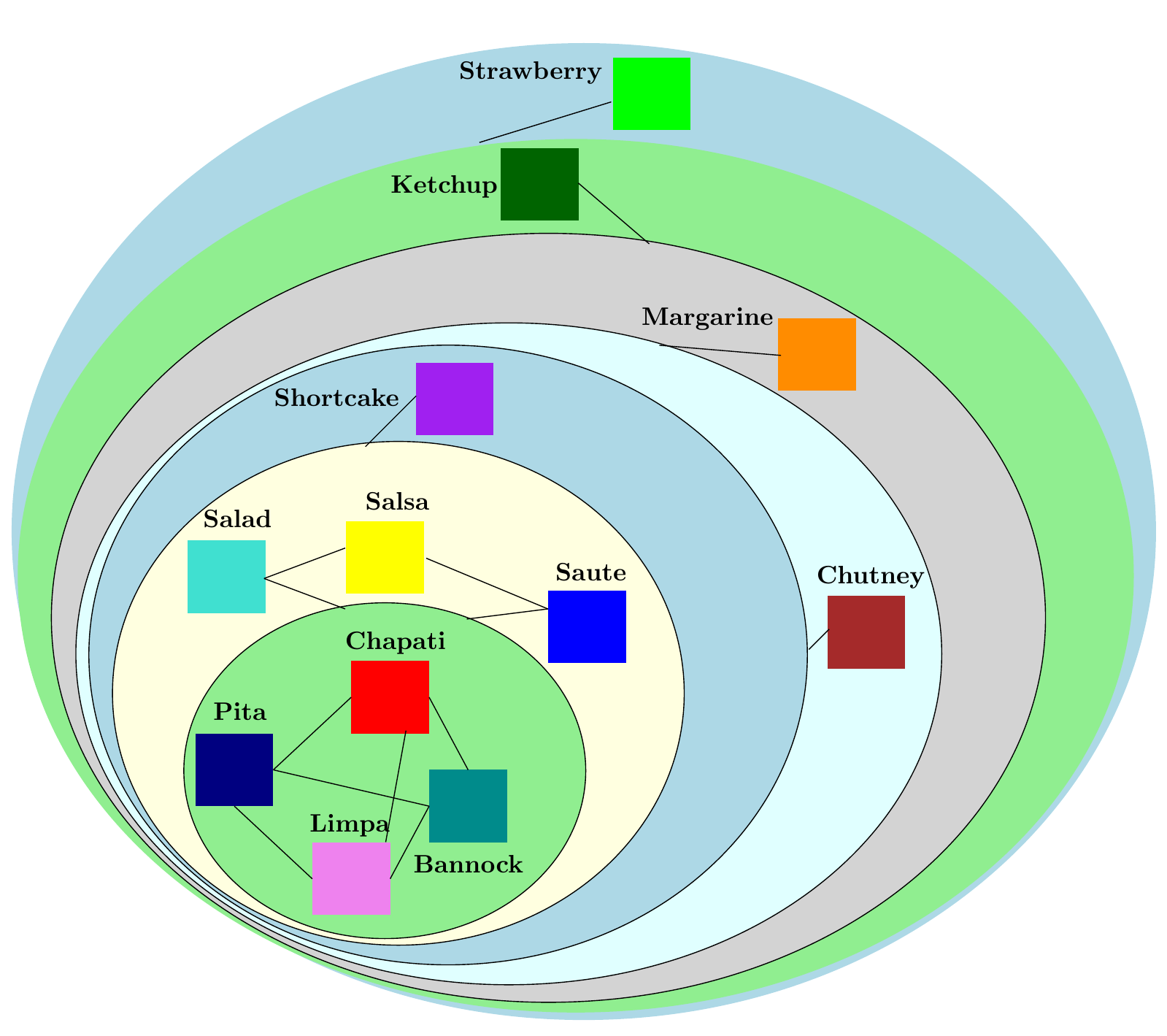}} 
\subfloat[{\bf $5^{th}$ order (conv$3$ layer reps.)}]{\includegraphics[width=58mm]{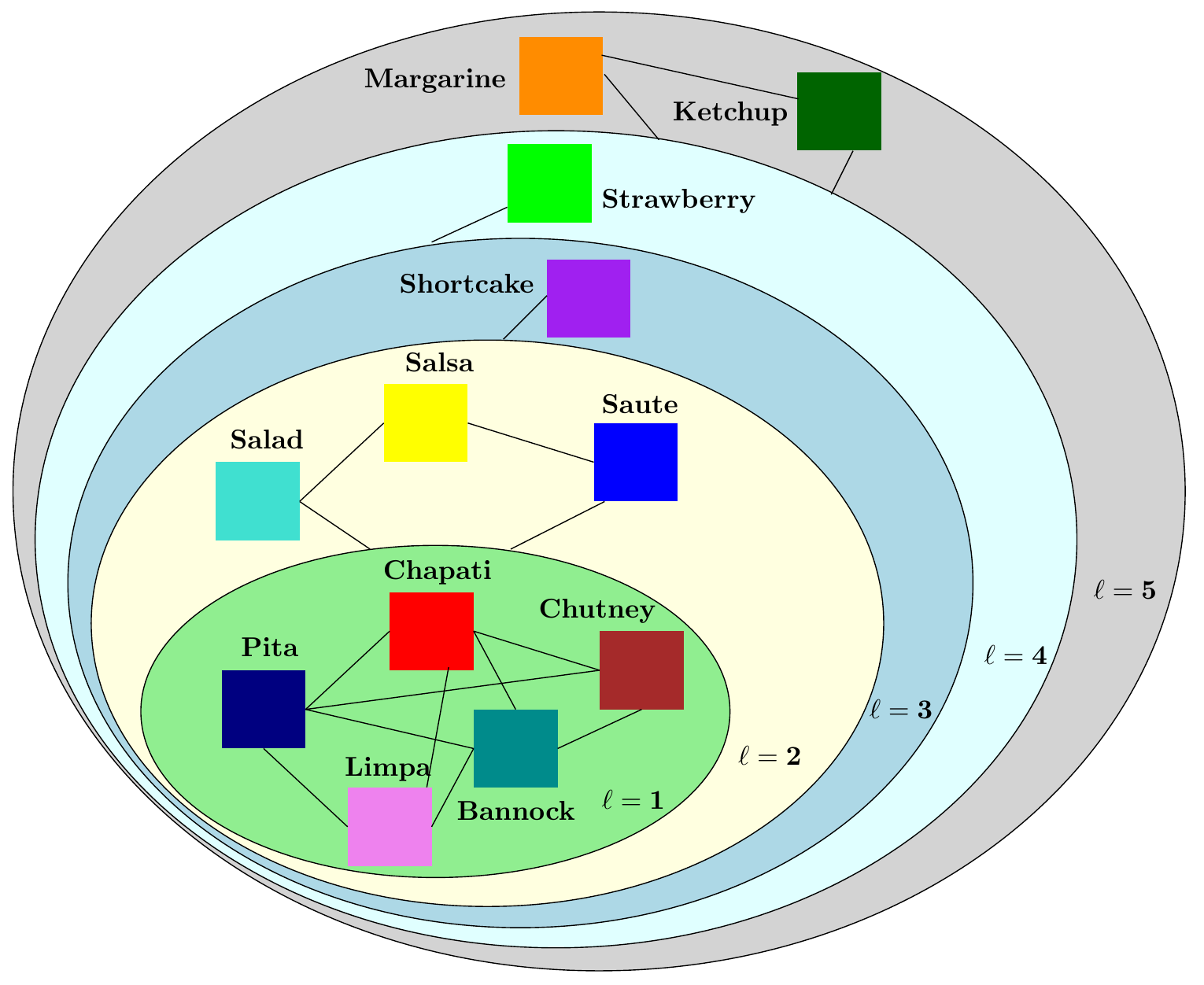}} 
\subfloat[{\bf $5^{th}$ order (Pixel reps.)}]{\includegraphics[width=58mm]{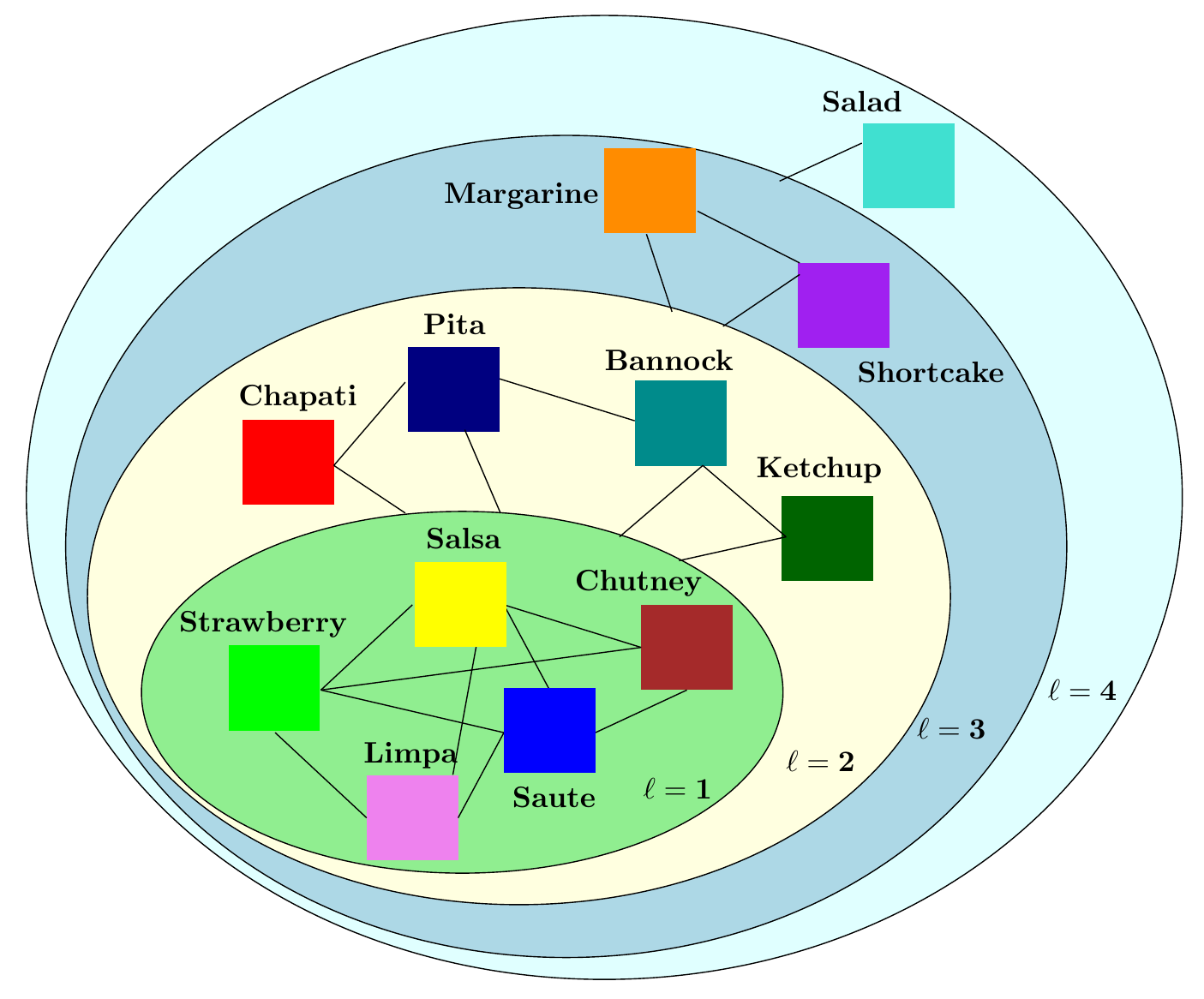}} 
\caption{\footnotesize \label{fig:food_vgg} {\bf Hierarchy and Compositions of VGG-S \cite{chatfield2014return} representations inferred by MMF.} 
(a) The $12$ classes, (b,c) Hierarchical structure from $5^{th}$ order MMF,
  (d) the structure from a $4^{th}$ order MMF, and (e,f) compositions from $3^{rd}$ conv. layer (VGG-S) and inputs.
$\tm = 0.1m$ (from Algorithm \ref{alg:multiolmmf}).}
\end{figure*}\mathversion{normal}


\subsection{MMF graphs} \label{sec:graphinterp}

The ability of a feature to succinctly represent the presence of an object/scene is, at least, in part, 
governed by the relationship of the learned representations across multiple object classes/categories. 
Beyond object-specific information, such cross-covariate contextual dependencies have shown to improve the performance 
in object tracking and recognition \cite{zhang2012robust} and medical applications \cite{ithapu2015imaging} (a motivating aspect of adversarial learning \cite{lowd2005adversarial}). 
Visualizing the histogram of gradients (HoG) features is one such interesting result that demonstrates the scenario where 
a correctly learned representation leads to a false positive \cite{vondrick2013hoggles}, for instance, the HoG features of a duck image are similar to a car HoG.
\cite{simonyan2013deep, dosovitskiy2015inverting} have addressed similar aspects for deep representations by visualizing image classification and detection models, 
and there is recent interest in designing tools for visualizing what the network perceives when predicting a test label \cite{yosinski2015understanding}. 
As shown in \cite{thebloglink}, the contextual images that a deep network (even with good detection power) desires to see may not even correspond to real-world scenarios.


The evidence from these works motivate a simple question -- {\it Do the semantic relationships learned by the deep representations associate with those seen by humans?} 
For instance, can such models infer that cats are closer to dogs than they are to bears;  
or that bread goes well with butter/cream rather than, say, salsa. 
Invariably, addressing these questions amounts to learning hierarchical and categorical relationships in the class-covariance of hidden representations.
Using classical techniques may not easily reveal interesting, human-relateable, trends as was shown very recently by \cite{peterson2016adapting}.
There are at least few reasons, but most importantly, the covariance of hidden representations (in general) has parsimonious structure with multiple compositions of blocks 
(the left two images in Figure \ref{fig:covs} are from AlexNet and VGG-S).
As motivated in Section \ref{sec:intro}, and later described in Section \ref{sec:mmfgraph} using Figure \ref{fig:example}, 
a MMF graph is the natural object to analyze such parsimonious structure.

\vspace{-2.5mm}
\subsubsection{Decoding the deep} \label{sec:decodedeep}

A direct application of MMF on the covariance of hidden representations reveals interesting hierarchical structure about the ``perception'' of deep networks.
To precisely walk through these compositions, consider the last hidden layer (FC$7$, that feeds into softmax) representations from a VGG-S network \cite{chatfield2014return} 
corresponding to $12$ different ImageNet classes, shown in Figure \ref{fig:food_vgg}(a).
Figure \ref{fig:food_vgg}(b,c) visualize a $5^{th}$ order MMF graph learned on this class covariance matrix.

\noindent {\bf The semantics of breads and sides.}
The $5^{th}$ order MMF says that the five categories -- {\it pita}, {\it limpa}, {\it chapati}, {\it chutney} and {\it bannock} -- 
are most representative of the localized structure in the covariance. 
Observe that these are four different flour-based main courses, and a side {\it chutney} that shared strongest context with the images of {\it chapati} in the training data 
(similar to the body building and dumbell images from \cite{thebloglink}).
MMF then picks {\it salad}, {\it salsa} and {\it saute} representations' at the $2^{nd}$ level, 
claiming that they relate the strongest to the {\it composition} of breads and {\it chutney} from the previous level (see visualization in Figure \ref{fig:food_vgg}(b,c)).
Observe that these are in fact the sides offered/served with bread.
Although VGG-S was $not$ trained to predict these relations, according to MMF, the representations are inherently learning them anyway -- a fascinating aspect of deep networks 
i.e., they are seeing what humans may infer about these classes. 


\noindent {\bf Any dressing? What are my dessert options?} 
Let us move to the $3^{rd}$ level in Figure \ref{fig:food_vgg}(b,c).
{\it margarine} is a cheese based dressing. {\it shortcake} is dessert-type meal made from {\it strawberry} (which shows up at $4^{th}$ level) and bread (the composition from previous levels). 
That is the full course. The last level corresponds to {\it ketchup}, which is an outlier, distinct from the rest of the $10$ classes -- 
a typical order of dishes involving the chosen breads and sides does not include hot sauce or ketchup.
Although {\it shortcake} is made up of strawberries, ``conditioned'' on the $1^{st}$ and $2^{nd}$ level dependencies, it is less useful in summarizing the covariance structure. 
An interesting summary of this hierarchy from  Figure \ref{fig:food_vgg}(b,c) is -- 
an order of {\it pita} with side {\it ketchup} or  {\it strawberries} is atypical in the data seen by these networks. 

\vspace{-2.5mm}
\subsubsection{Are we reading tea leaves?} \label{sec:tealeaves}

It is reasonable to ask if this description is meaningful since the semantics drawn above are subjective.
We provide explanations below. 
First, the networks are $not$ trained to learn the hierarchy of categories -- the task was object/class detection.
Hence, the relationships are completely a by-product of the power of deep networks to learn contextual information, 
{\it and} the ability of MMF to model these compositions by uncovering the structure in the covariance matrix. 
Supplement provides further evidence by visualizing such hierarchy from few dozens of other ImageNet classes. 
Second, one may ask if the compositions are sensitive/stable to the order $k$ -- a critical hyperparameter of MMF. 
Figure \ref{fig:food_vgg}(d) uses a $4^{th}$ order MMF, and the resulting hierarchy is similar to that from Figure \ref{fig:food_vgg}(b).
Specifically, the different breads and sides show up early, and the most distinct categories ({\it strawberry} and {\it ketchup}) appear at the higher levels.
Similar patterns are seen for other choices of $k$ (see supplement).

Further, if the class hierarchy in Figures \ref{fig:food_vgg}(b--d) is non-spurious, then similar trends should be implied by MMF's on different (higher) layers of VGG-S.
Figure \ref{fig:food_vgg}(e) shows the compositions from the $10^{th}$ layer representations (the outputs from $3^{rd}$ convolutional layer of VGG-S) of the $12$ classes in Figure \ref{fig:food_vgg}(a).
The strongest compositions, the $8$ classes from $\ell=1$ and $2$, are already picked up half-way thorough the VGG-S, 
providing further evidence that the compositional structure implied by MMF is data-driven. We further discuss this in Section \ref{sec:mmfflow}.
Finally, we compared MMF's class-compositions to the hierarchical clusters obtained from agglomerative clustering of representations.
The relationships in Figure \ref{fig:food_vgg}(b-d) are not apparent in the corresponding dendrograms (see supplement, \cite{peterson2016adapting}) -- 
for instance, the dependency of {\it chutney}/{\it salsa}/{\it salad} on several breads, or the disparity of {\it ketchup} from the others. 

Overall, Figure \ref{fig:food_vgg}(b--e) shows many of the summaries that a human may infer about the $12$ classes in Figure \ref{fig:food_vgg}(a). 
Apart from visualizing deep representations, such MMF graphs are vital exploratory tools for category/scene understanding from unlabeled representations 
in transfer and multi-domain learning \cite{ben2010theory}.
This is because, by comparing the MMF graph {\it prior} to inserting the new unlabeled instance to the one {\it after} insertion, 
one can infer whether the new instance contains non-trivial information that cannot be expressed as a composition of existing categories.

\vspace{-3.5mm}
\subsubsection{The flow of MMF graphs: An exploratory tool} \label{sec:mmfflow}

Figure \ref{fig:food_vgg}(f) shows the compositions from the $5^{th}$ order MMF on the {\it input} (pixel-level) data. 
These features are non-informative, and clearly, the classes whose RGB values correlate are at $l=0$ in Figure \ref{fig:food_vgg}(f).
But most importantly, comparing Figure \ref{fig:food_vgg}(b,e) we see that $l=1$ and $2$ have the same compositions. 
One can construct visualizations like Figure \ref{fig:food_vgg}(b,e,f) for all the layers of the network.
Using this trajectory of the class compositions, one can ask whether a new layer needs to be added to the network (a vital aspect for model selection in deep networks \cite{ithapu2017architectural}).
This is driven by the saturation of the compositions -- if the last few levels' hierarchies are similar, then the network has already learned the information in the data.
On the other hand, variance in the last levels of MMFs implies that adding another network layer may be beneficial. 
The saturation at $l=1$, $2$ in Figure \ref{fig:food_vgg}(b,e) (see supplement for remaining layers' MMFs) is one such example. 
If these $8$ classes are a priority, then the predictions of the VGG-S' $3^{rd}$ convolutional layer may already be {\it good enough}.
Such constructs can be tested across other layers and architectures (see project webpage).


\section{Conclusions}

We present an algorithm that uncovers multiscale structure of symmetric matrices by performing a matrix factorization. 
We showed that it is an efficient importance sampler for relative leveraging of features.
We also showed how the factorization sheds light on the semantics of categorical relationships encoded in deep networks, 
and presented ideas to facilitate adapting/modifying their architectures. 

{\small \noindent {\bf Acknowledgments:} The authors are supported by NIH AG021155, EB022883, AG040396, NSF CAREER 1252725, NSF AI117924 and 1320344/1320755.}


{\small
\bibliographystyle{ieee}
\bibliography{highordermmfARXIV}
}
\end{document}